\documentclass[conference]{IEEEtran}
\IEEEoverridecommandlockouts
\usepackage{cite}
\usepackage{amsmath,amssymb,amsfonts}
\usepackage{algorithmic}
\usepackage{graphicx}
\usepackage{textcomp}
\usepackage{xcolor}
\usepackage[hidelinks]{hyperref}
\usepackage{caption}
\usepackage{subcaption}
\usepackage{enumitem}
\usepackage{url}  % For handling URLs

\def\BibTeX{{\rm B\kern-.05em{\sc i\kern-.025em b}\kern-.08em
    T\kern-.1667em\lower.7ex\hbox{E}\kern-.125emX}}

\begin{document}

% \title{Self Supervised Learning for Wildfire Detection}
\title{Generative AI for Enhanced Wildfire Detection: Bridging the Synthetic-Real Domain Gap}

\author{\IEEEauthorblockN{Satyam Gaba}
\IEEEauthorblockA{\textit{Univeristy of the Cumberlands} \\
\textit{Williamsburg, US} \\
sgaba27353@ucumberlands.edu}
% \and
% \IEEEauthorblockN{2\textsuperscript{nd} David Ostrowski}
% \IEEEauthorblockA{\textit{School Of Professional Studies}\\ \textit{Northwestern University} \\
% \textit{Evanston, US}\\
% david.ostrowski@northwestern.edu}
% \and
% \IEEEauthorblockN{3\textsuperscript{rd} Given Name Surname}
% \IEEEauthorblockA{\textit{dept. name of organization (of Aff.)} \\
% \textit{name of organization (of Aff.)}\\
% City, Country \\
% email address or ORCID}
}

\maketitle

\begin{abstract}
The early detection of wildfires is a critical environmental challenge, with timely identification of smoke plumes being key to mitigating large-scale damage. While deep neural networks have proven highly effective for localization tasks, the scarcity of large, annotated datasets for smoke detection limits their potential.

In response, we leverage generative AI techniques to address this data limitation by synthesizing a comprehensive, annotated smoke dataset. We then explore unsupervised domain adaptation methods for smoke plume segmentation, analyzing their effectiveness in closing the gap between synthetic and real-world data.

To further refine performance, we integrate advanced generative approaches such as style transfer, Generative Adversarial Networks (GANs), and image matting. These methods aim to enhance the realism of synthetic data and bridge the domain disparity, paving the way for more accurate and scalable wildfire detection models.
\end{abstract}

\section{Introduction}

Wildfires pose a significant threat in many regions of the world, particularly those with hot climates and vast areas of vegetation. Traditional methods of wildfire detection, relying on human observers, are both labor-intensive and time-consuming. To address this, automated wildfire detection systems have been developed, utilizing infrared sensors to detect heat from fires and light detection and ranging (LiDAR) systems to monitor smoke particles. However, these systems can be costly and prone to false alarms due to atmospheric interference. To improve reliability and coverage, more advanced technologies such as CCD cameras have been deployed. One such initiative, ALERTCalifornia project \cite{alertcalifornia}, was launched to enhance wildfire monitoring by using cameras to capture fire and smoke activity. The system’s web-based interface helps pinpoint the exact location of new fires, significantly reducing the time needed to confirm fire ignition and dispatch resources, such as engines or aircraft, to the scene.

\begin{figure*}
\centering
\begin{subfigure}{.20\textwidth}
    \centering
    \includegraphics[width=0.9\linewidth]{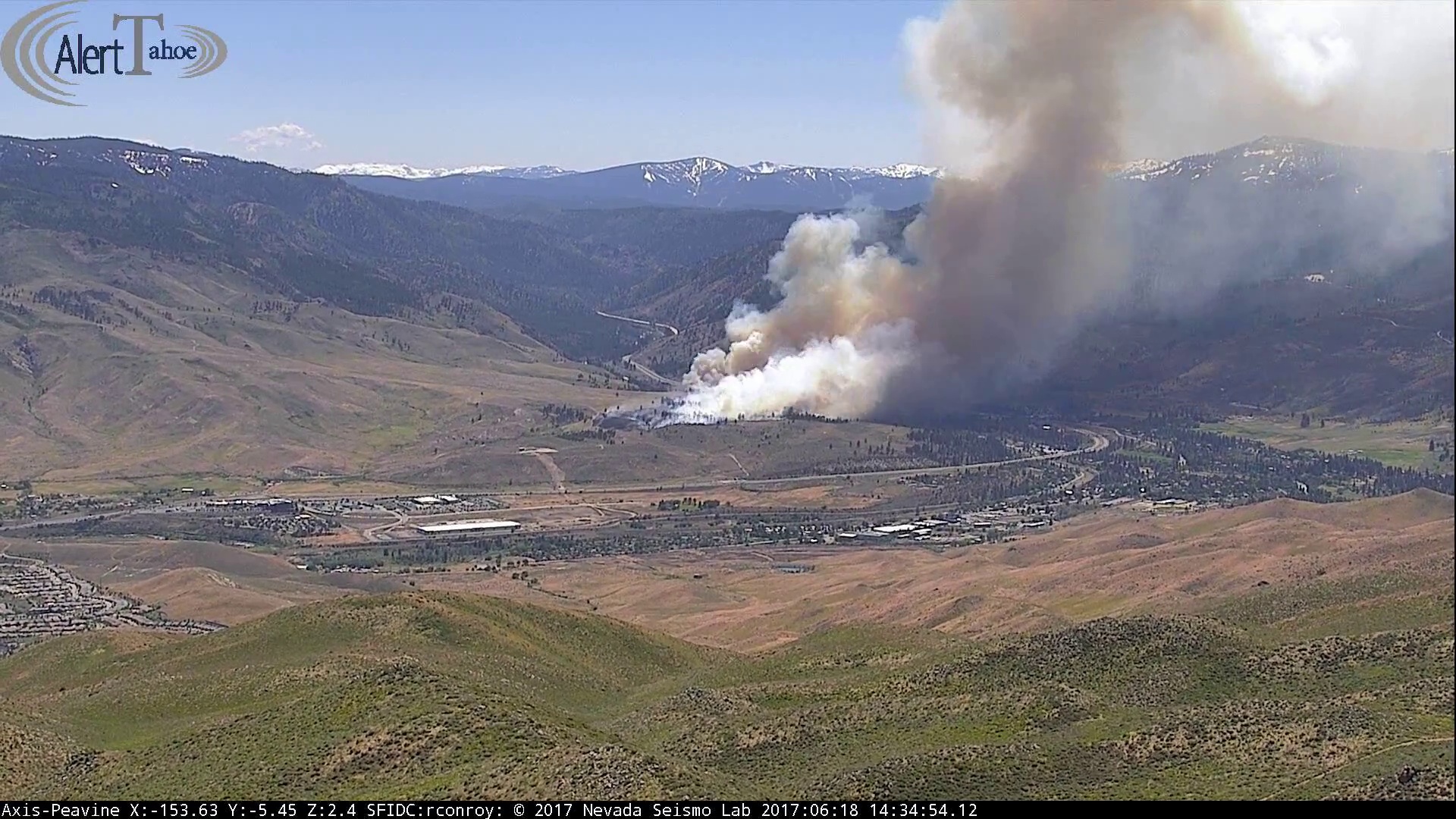}
    % \label{advent_res_a}
    \caption{}
\end{subfigure}%
\begin{subfigure}{.20\textwidth}
    \centering
    \includegraphics[width=0.9\linewidth]{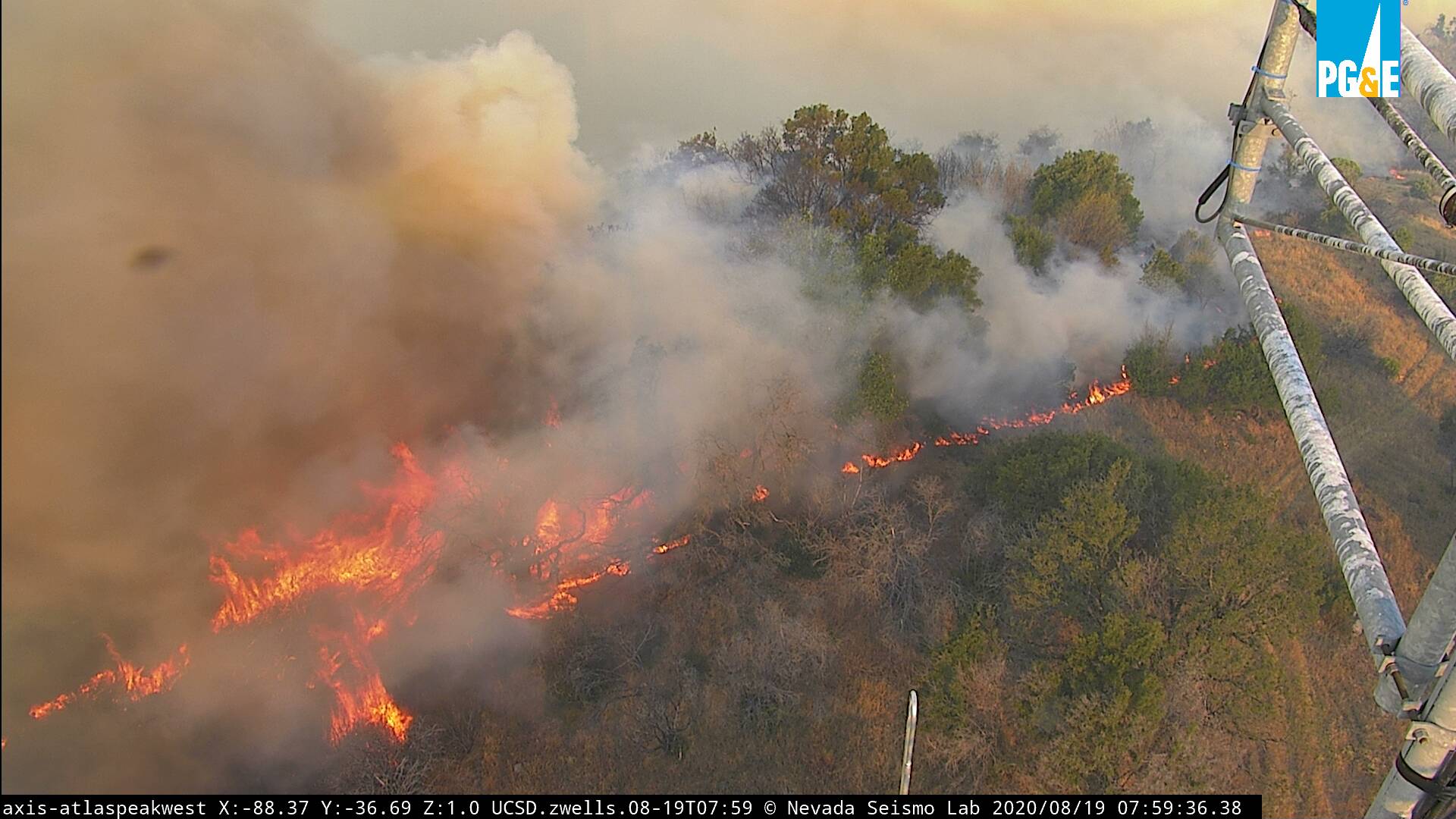}
    % \label{}
    \caption{}
\end{subfigure}%
\begin{subfigure}{.20\textwidth}
    \centering
    \includegraphics[width=0.9\linewidth]{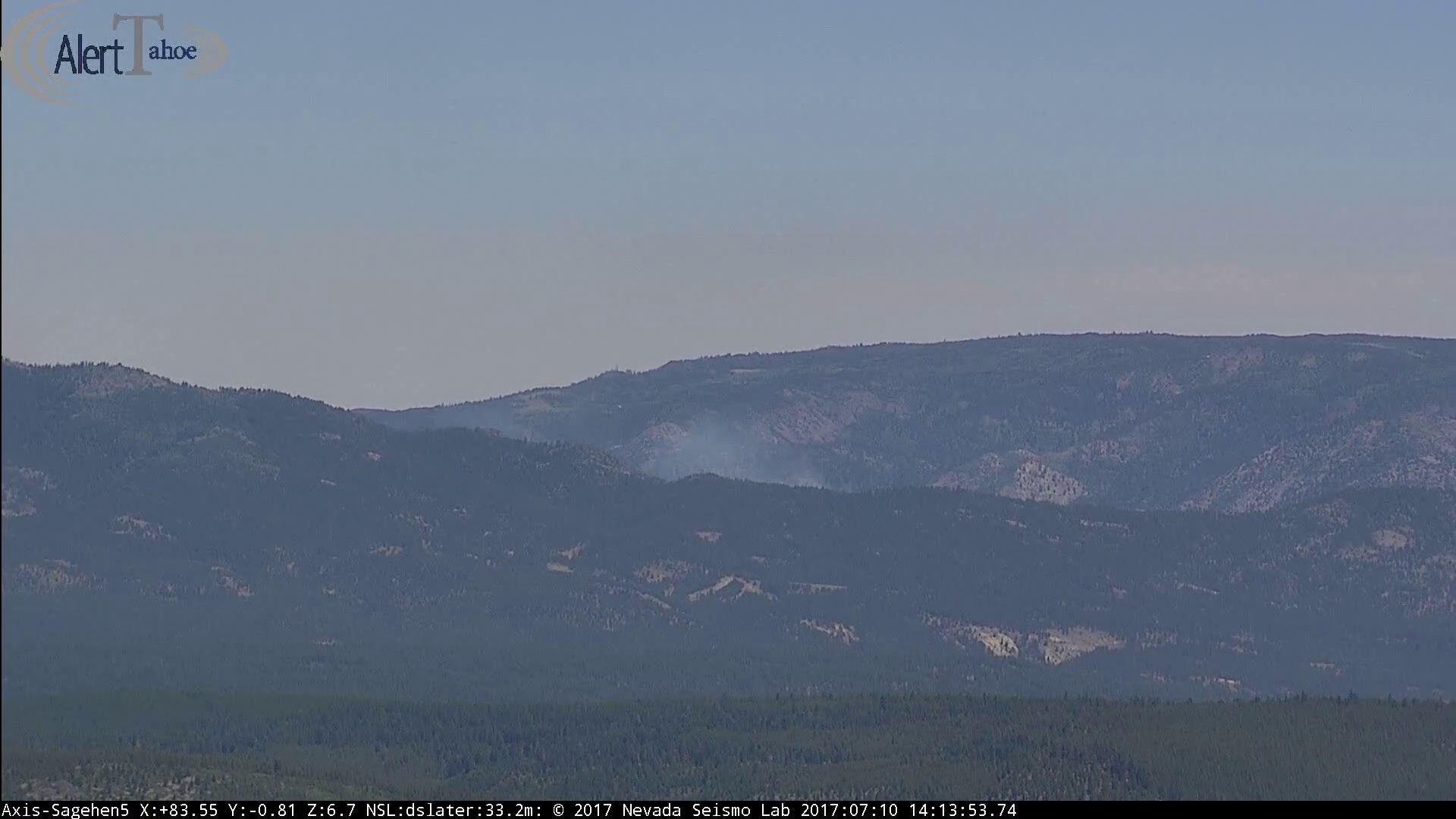}
    % \label{}
    \caption{}
\end{subfigure}%
\begin{subfigure}{.20\textwidth}
    \centering
    \includegraphics[width=0.9\linewidth]{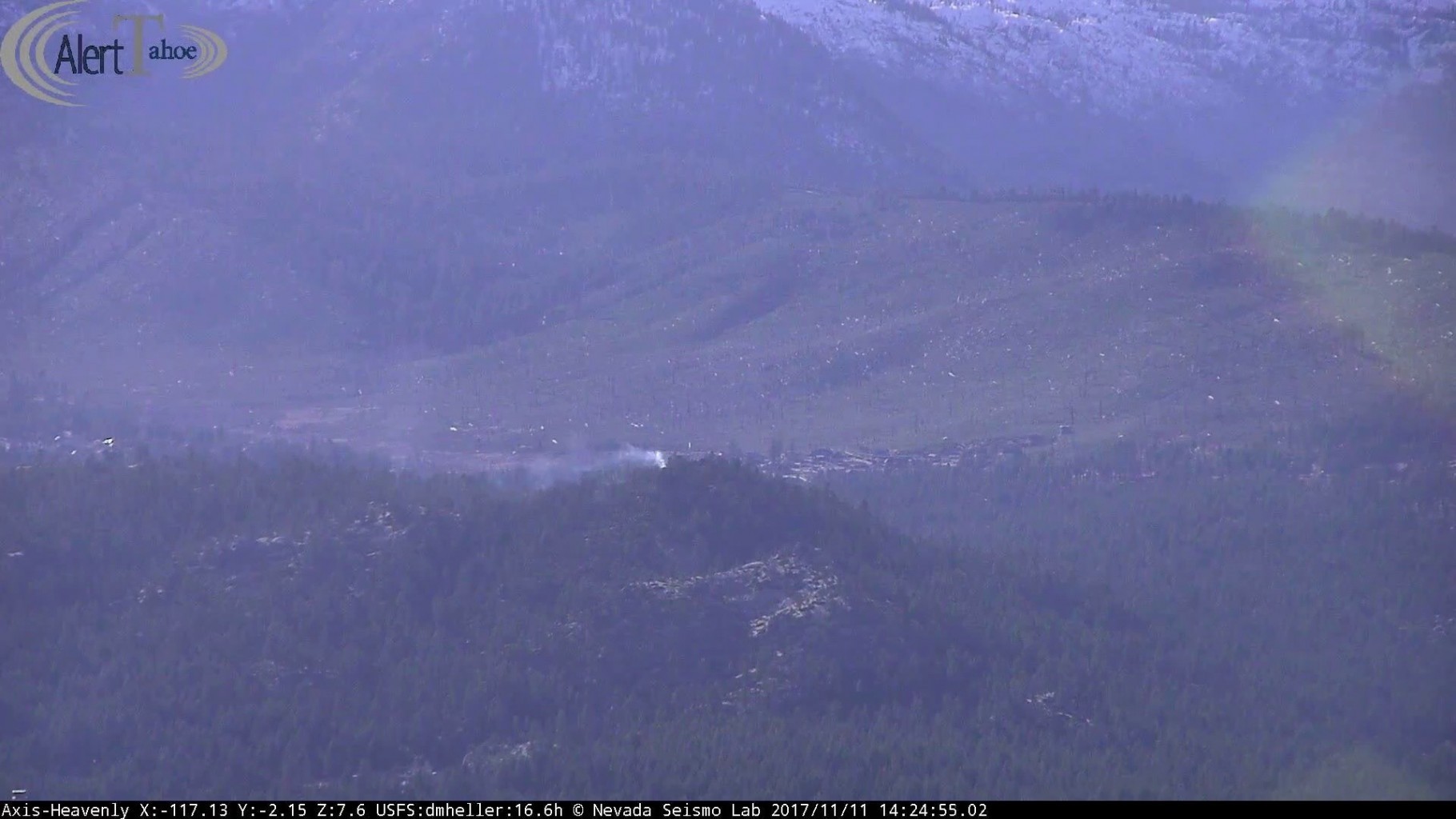}
    % \label{}
    \caption{}
\end{subfigure}%
\begin{subfigure}{.20\textwidth}
    \centering
    \includegraphics[width=0.9\linewidth]{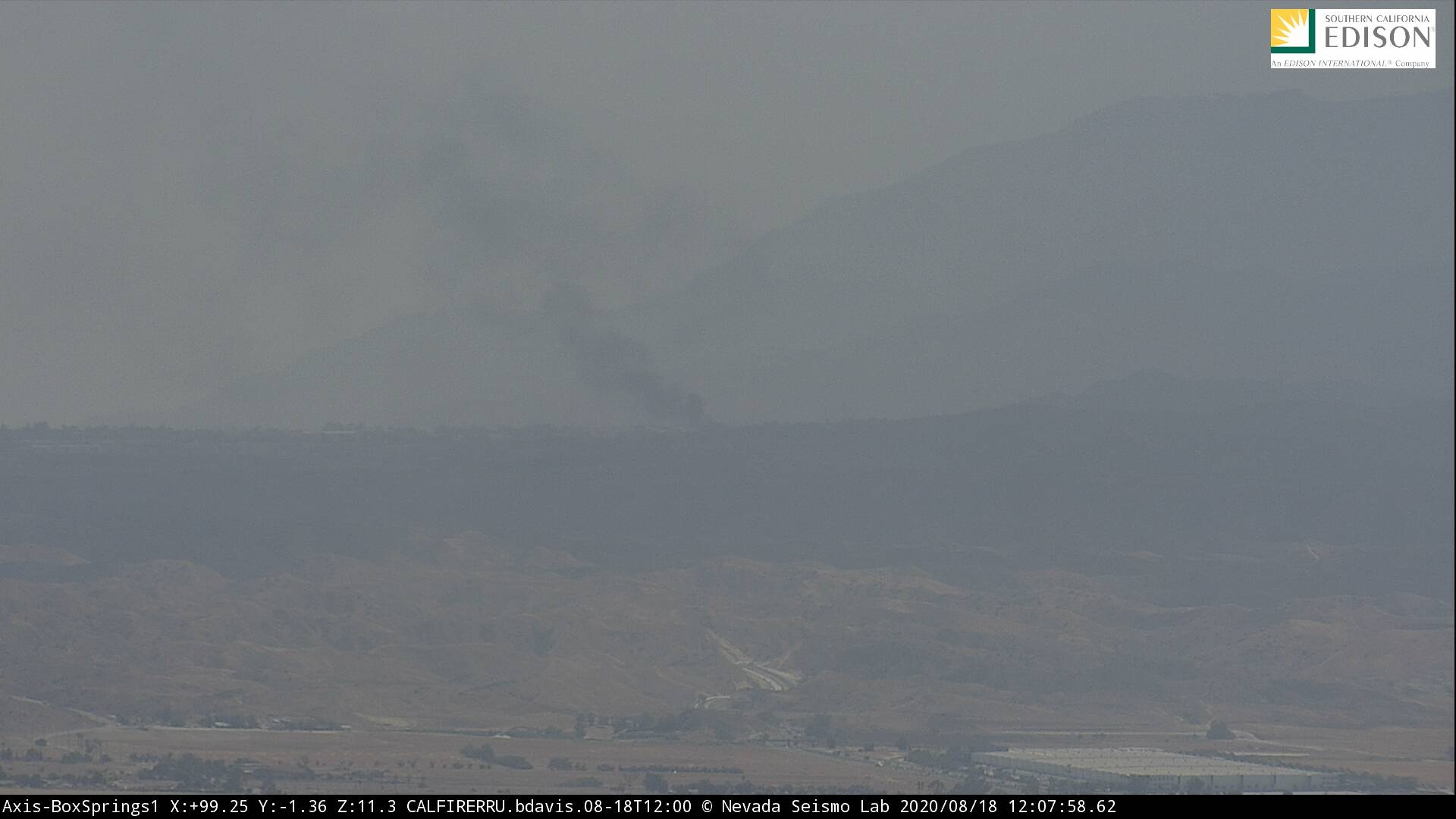}
    % \label{}
    \caption{}
\end{subfigure}%
\caption{Examples of real wildfire images captured by cameras deployed for the ALERTCalifornia Project. The variety in these images introduces significant challenges for the segmentation task. Images are taken at different times, from various cameras positioned in different locations. Challenges include occlusions (b), small or distant smoke plumes (c), low lighting conditions (d), and haze caused by smoke (e).}
\label{fig:smoke}
\end{figure*}

Detecting wildfire smoke is a difficult task because the main characteristics of wildfire smoke are vague color patterns, low spreading speed, and indistinct shape. It's non-rigid and deformable nature might also result in a loss of details. This is clear from (Fig.\ref{fig:smoke}) where the images have other interferences in the form of smog, haze and clouds that merge with the smoke plumes and have color and textural characteristics similar to those of smoke.

Feiniu et al. \cite{yaun18deepsmokesegmentation} classify smoke detection into two categories: whole image smoke recognition and smoke detection/segmentation. Whole image recognition identifies smoke presence without locating it, while smoke detection focuses on determining its position, often framed as an object detection task using bounding boxes. However, the diffuse, irregular nature of smoke makes defining accurate boundaries challenging. Smoke segmentation, by contrast, provides pixel-level precision, generating fine-grained masks to separate smoke from the background. While more complex, segmentation achieves higher accuracy by clearly distinguishing the smoke foreground from its background.

To detect and localize wildfire smoke, we generate semantic masks of the smoke within an image. However, due to the lack of labeled smoke masks and prohibitive cost of labeling required for training a semantic segmentation network, we employ Unsupervised Domain Adaptation (UDA) techniques to predict these masks. For UDA, we create synthetic smoke masks and overlay them onto background (non-smoke) images, as detailed in Section \ref{sec:syn_data}. In this setup, the synthetic images serve as the source domain, while real smoke images represent the target domain.

We conducted experiments using two state-of-the-art Unsupervised Domain Adaptation (UDA) semantic segmentation models, AdaptSegNet and AdvEnt, trained on synthetic smoke images and evaluated on real wildfire images. The results highlight a significant domain gap between the source (synthetic smoke) and target (real wildfire) domains, rendering direct use of synthetic data impractical. To reduce this gap, we focused on generating synthetic images resembling real-world scenarios. We hypothesize reasons behind this gap and explore mitigation approaches in Section \ref{sec:bridge}. Using Generative Adversarial Networks (GANs), we tested Style Transfer, Pix2Pix GAN, and Cycle GAN for pixel-level domain alignment, but results show challenges due to the deformable, dynamic nature of smoke, which complicates segmentation and transfers background features between domains. To address the lack of precise smoke masks, we further experimented with Deep Image Matting (Section \ref{sec:img_matting}), yielding promising results. Detailed outcomes are provided in Section \ref{sec:exp}.

\section{Related Work}

\subsection{Traditional methods}
Several studies have employed various traditional techniques for wildfire smoke detection. Some approaches utilize temporal features combined with heuristic parameters to identify smoke, while others incorporate spatiotemporal characteristics along with random forest classifiers to enhance detection accuracy. Additionally, Gaussian Mixture Models (GMM) have been applied to smoke segmentation, as demonstrated in \cite{gauss_mix}. More recently, deep learning-based frameworks such as Fully Convolutional Networks (FCN) have been explored for smoke segmentation tasks, as discussed in \cite{yaun18deepsmokesegmentation}.

\subsection{Smoke Segmentation networks}
A notable recent work on smoke segmentation using Fully Convolutional Networks (FCNs) has been conducted by Yuan et al. \cite{yaun18deepsmokesegmentation}, who developed a two-model network specifically for smoke segmentation. Their approach relies on synthetic data for training and is evaluated on real-world smoke images. However, their performance on real-world data, particularly on the wildfire smoke images from the ALERTCalifornia network data, has been limited. Other segmentation networks, such as UNet \cite{ronneberger2015unet}, have also been applied to smoke segmentation but have shown similar performance challenges.

\subsection{Domain Adaptation for Semantic Segmentation}
In recent years, significant progress has been made in both supervised and unsupervised domain adaptation (UDA) techniques for semantic segmentation. Our work focuses on UDA methods, which aim to bridge the domain gap without relying on labeled target domain data.

A common approach to UDA involves adversarial training, where features are adapted across domains using a domain discriminator. Hoffman et al. \cite{hoffman2016fcns} pioneered adversarial-based domain adaptation for semantic segmentation, utilizing a domain discriminator that operates on intermediate feature representations from the segmentation network. Their method also includes a per-image label distribution loss to enforce scene layout similarity and semantic segmentation error on labeled source data. Chen et al. \cite{chen2018road} extended this work by incorporating a target-guided distillation loss to learn the target domain image style, reducing the risk of overfitting on source data.

Zhang et al. \cite{zhang2018fully} introduced a dual-network approach, where one network generates images in the style of the other domain, and the second network predicts semantic segmentation for both domains. A domain discriminator is applied to segmentation outputs, while Atrous Spatial Pyramid Pooling is used to capture multi-scale features, improving domain differentiation.

Other works, such as CyCada \cite{hoffman2017cycada}, CrDoCo \cite{chen2020crdoco}, and Toldo et al. \cite{Toldo_2020}, leverage Generative Adversarial Networks (GANs) for domain adaptation. Recent approaches, including Tsai et al. \cite{tsai2020learning}, Chen et al. \cite{chen2019learning}, Luo et al. \cite{luo2019taking}, and Biasetton et al. \cite{biasetton2019unsupervised}, focus on adaptation from the segmentation outputs rather than intermediate feature representations. Tsai et al. \cite{tsai2020learning} utilize a domain discriminator that classifies the source of semantic segmentation predictions at multiple levels, adversarially training both the domain discriminator and the segmentation model. In contrast, researchers such as Vu et al. \cite{vu2019advent} explore entropy minimization techniques as an alternative to adversarial training for feature adaptation. Additionally, recent advancements in Generative AI, particularly for image generation and authentication, have further demonstrated the versatility of these approaches in related domains \cite{Nandwani2024}.

\subsection{Smoke Segmentation using Domain Adaptation}
Xu et al. \cite{xu2018domain} were among the first to propose a domain adaptation model specifically for smoke detection and localization. Their approach utilizes Single Shot Detector (SSD) and Multi-Scale Convolutional Neural Network (MS-CNN) for smoke localization, leveraging Maximum Mean Discrepancy (MMD) and Correlation Alignment (CORAL) loss \cite{sun2016deep} to minimize domain discrepancy in a semi-supervised setting. These losses help adapt features learned from synthetic data to real-world smoke data.

Their method focuses on object detection rather than semantic segmentation, which is a notable limitation. The inherent challenges of smoke's deformable nature and the difficulty in generating precise masks, as mentioned earlier, restrict the applicability of semantic segmentation for this task. Although they report some quantitative results on smoke localization accuracies using their custom dataset, the lack of publicly available code and an unclear explanation of their training strategy make it difficult to replicate their results or extend the model to other datasets.

\section{Method}
Since obtaining labeled real-world smoke data is extremely challenging, and annotating new datasets is both prohibitively expensive and time-consuming, it becomes difficult to efficiently train deep convolutional neural networks (CNNs) that require large, well-annotated datasets. To address this issue, we generate a synthetic smoke dataset by overlaying smoke plumes extracted from real smoke images onto non-smoke backgrounds. The process for creating this synthetic dataset is detailed in Section \ref{sec:syn_data}. This approach allows us to circumvent the limitations of real-world data availability while still providing the necessary diversity and volume for effective model training.

We initially train our smoke segmentation model on the large synthetic wildfire dataset and observe poor performance of segmentation models due to the significant domain gap between synthetic and real-world data. To address this, we investigate several domain adaptation techniques aimed at adapting a model trained on the synthetic (source) domain to perform effectively on real-world (target) data. These techniques are described in detail in Section \ref{sec:uda}.

Our results reveal that the gap between the synthetic and real domains is substantial, and even state-of-the-art domain adaptation techniques struggle to bridge this gap effectively. To mitigate this large domain discrepancy, we explore multiple approaches, including Style Transfer, Pix2Pix GAN, CycleGAN, and Image Matting, which are discussed in detail in Section \ref{sec:bridge}.

\subsection{Synthetic Data Generation}
\label{sec:syn_data}
For synthetic smoke data generation, we build upon the techniques used by Yuan et al. \cite{yaun18deepsmokesegmentation} and Xu et al. \cite{Xu_2017}, where smoke frames are blended with non-smoke backgrounds. Unlike these methods that rely on alpha channels generated by renderers for effective blending, we introduce a more flexible approach by utilizing the intensity values of both the smoke images and non-smoke backgrounds as proxies for alpha channels. This technique allows us to blend the smoke and background more effectively and makes it possible to use any RGB smoke image as a source for our synthetic data.

Furthermore, we introduce significant variation in smoke size, shape, direction, color, and transparency, enabling the creation of a large, diverse dataset of realistic smoke using a relatively small set of smoke and background images. 

\begin{figure}[htbp]
    \centering
    \begin{subfigure}[t]{0.5\columnwidth}
        \includegraphics[width=\columnwidth]{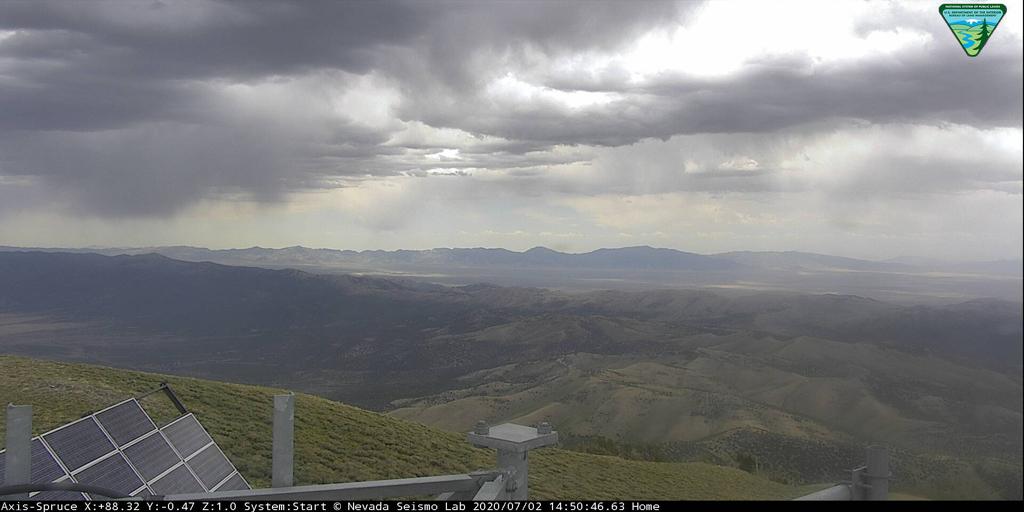}
        \caption{Background}
    \end{subfigure}%
    ~
    \begin{subfigure}[t]{0.5\columnwidth}
        \includegraphics[width=\columnwidth]{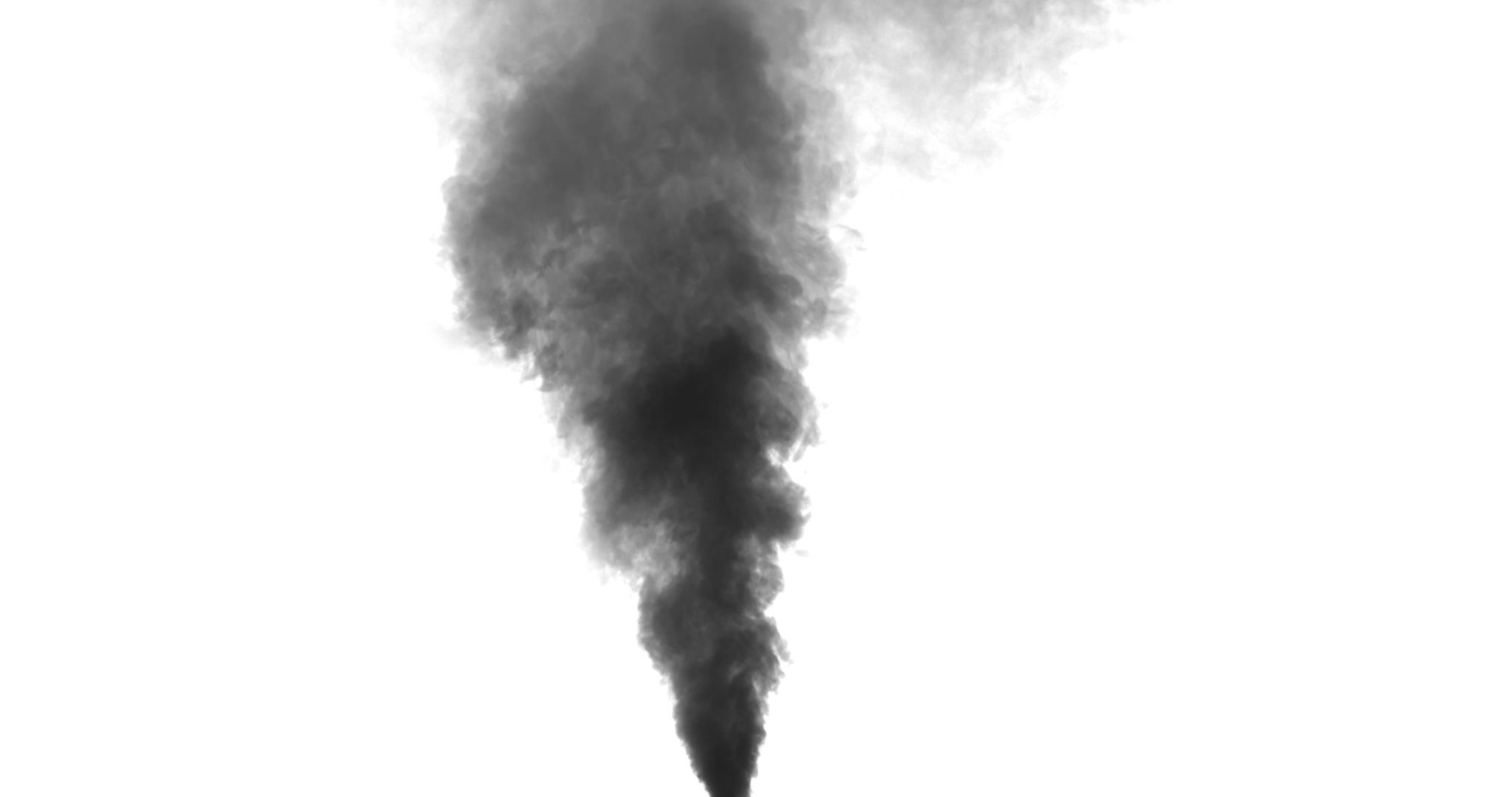}
        \caption{Smoke to be overlayed}
    \end{subfigure}
    
    \begin{subfigure}[t]{0.5\columnwidth }
        \includegraphics[width=\columnwidth]{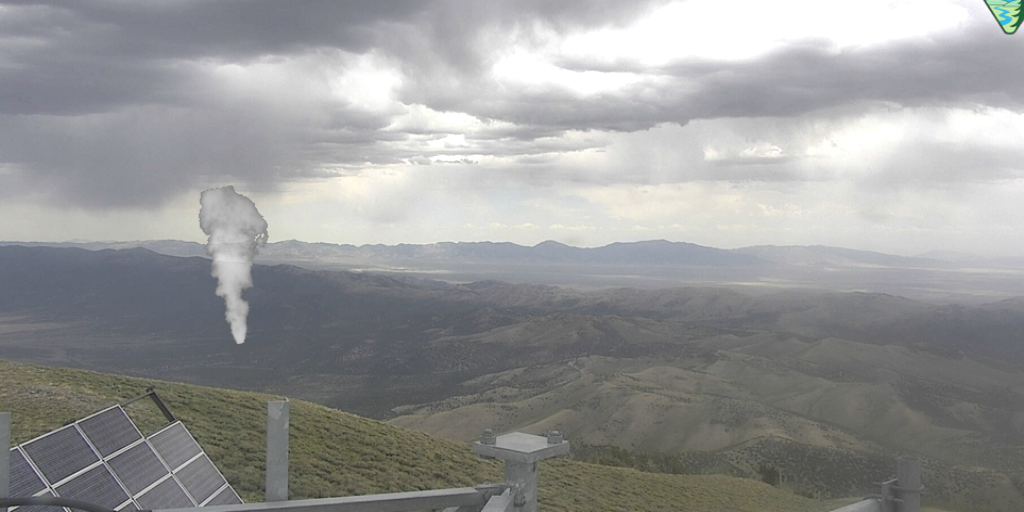}
        \caption{Generated Synthetic image}
    \end{subfigure}%
    ~
    \begin{subfigure}[t]{0.5\columnwidth}
        \includegraphics[width=\columnwidth]{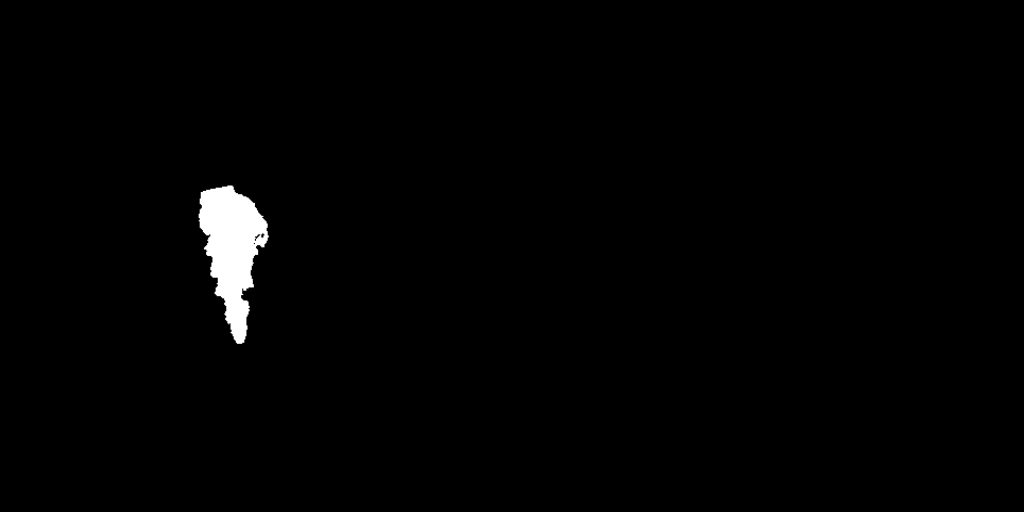}
        \caption{Corresponding Segmentation Mask}
    \end{subfigure}
    \caption{Example of synthetically generated data from smoke and non-smoke images}
    \label{fig:synth_data}
\end{figure}

As our work focuses on smoke segmentation using data from the Alert Wildfire camera network, we use non-smoke images from this network as the background onto which smoke is overlaid. Figure \ref{fig:synth_data} shows an example of synthetic data generated by overlaying an RGB frame of smoke, sourced from the internet and captured against a white background, onto one of the non-smoke images from ALERTCalifornia camera network.

For more details on the data generation process, please see our code. 

\subsection{Transfer learning}
Due to the limited availability of labeled real-world data, training an effective segmentation model directly for the target domain is challenging. To address this, we employ transfer learning as a means to estimate an upper bound on the model's performance in the target domain.

In this approach, we fine-tune our U-Net model, which was pretrained on synthetic data, using a subset of labeled real-world target data. Initially, the U-Net segmentation model is trained on the synthetic dataset, after which it is fine-tuned on real wildfire images through transfer learning. To optimize performance, we vary the amount of labeled target data used for fine-tuning and select the model that achieves satisfactory results with the minimal amount of labeled real data. This method helps to maximize the utility of the limited labeled dataset while improving model performance on real-world wildfire images.

\subsection{Unsupervised Domain Adaptation}
\label{sec:uda}
We lack sufficient labeled real-world smoke data, so our approach focuses on learning the semantics of smoke from synthetic data and adapting it to real-world smoke through Unsupervised Domain Adaptation (UDA) techniques. The UDA methods we explore for our segmentation task are detailed in Sections \ref{sec:asn} and \ref{sec:advent}. In UDA, the source domain is assumed to be well-labeled, while the target domain remains entirely unlabeled. For our experiments, we use labeled synthetic smoke images as the source domain and real-world smoke images as the target domain. The goal is to train a network that performs effectively on the target domain.

\begin{figure}[htbp]
\centering
    \begin{subfigure}[t]{0.5\columnwidth}
        \includegraphics[width=0.9\columnwidth]{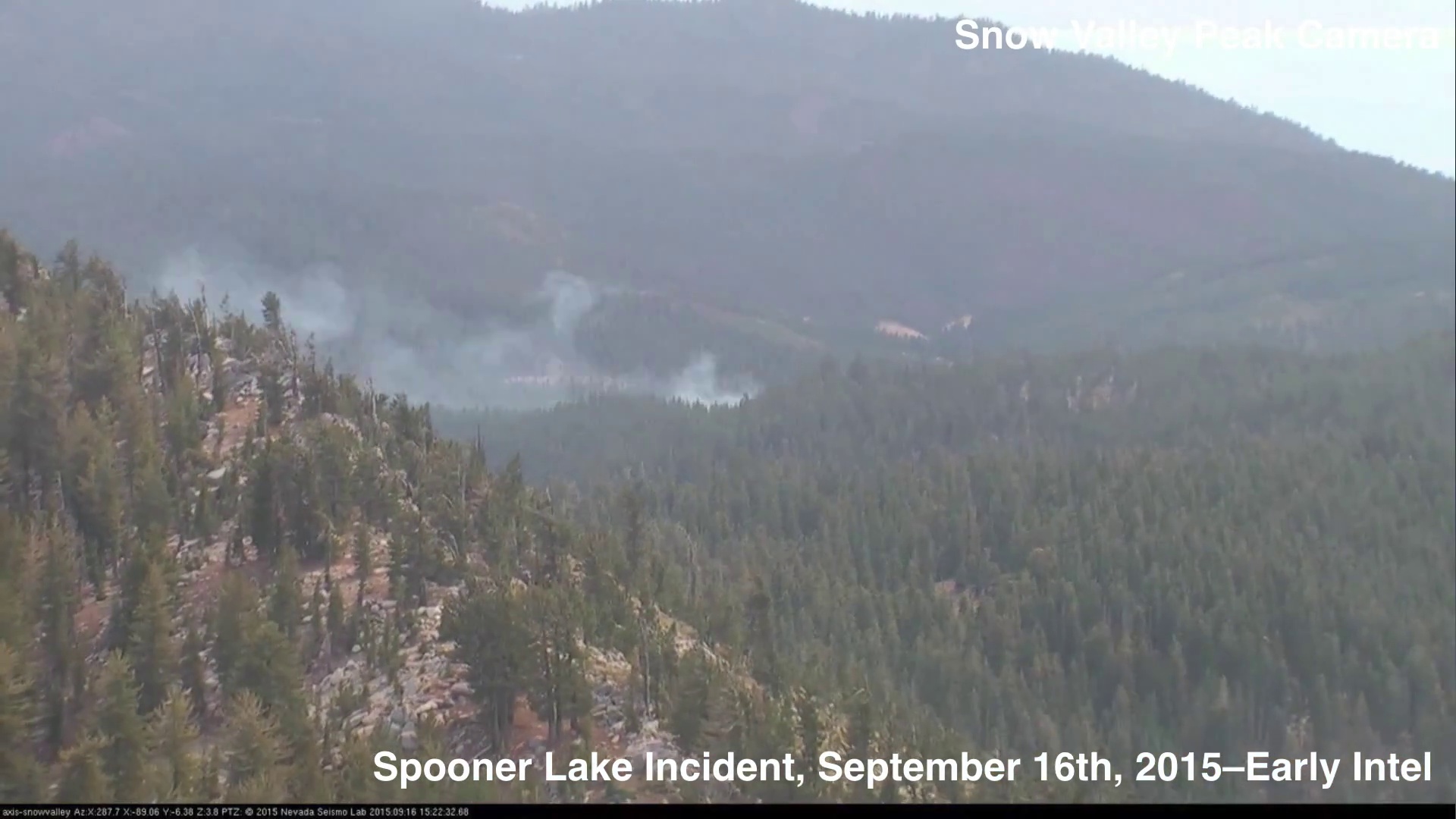}
        \caption{RGB Image}
    \end{subfigure}%
        \begin{subfigure}[t]{0.5\columnwidth}
        \includegraphics[width=0.9\columnwidth]{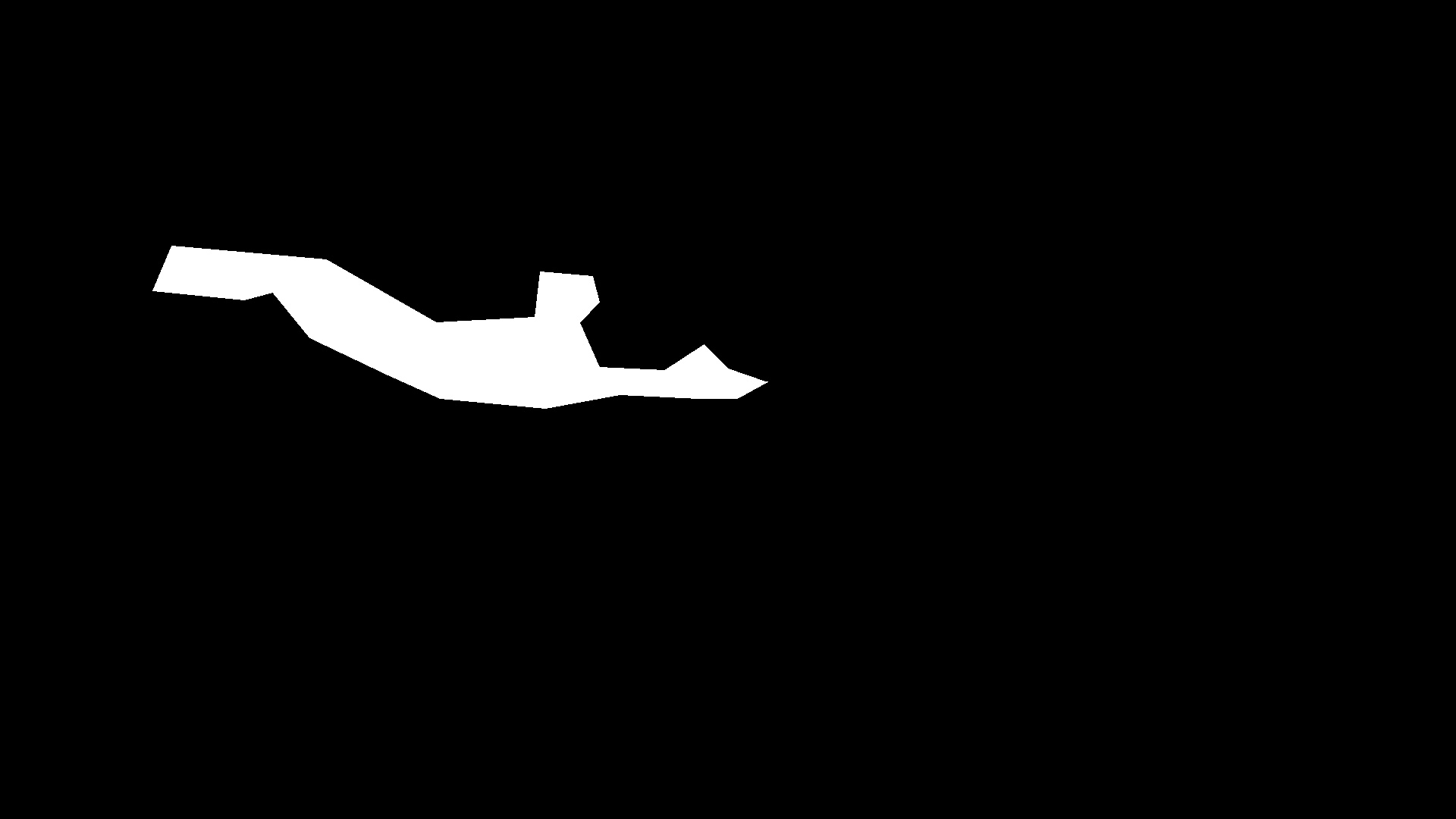}
        \caption{Labelled Mask}
    \end{subfigure}
\caption{Manually annotated smoke data using VGG Annotation tool \cite{vgg_ann}.}
\label{fig:labeling}
\end{figure}

To evaluate the performance of UDA techniques on real smoke images, we manually annotated 400 real smoke images using the VGG Annotator tool \cite{vgg_ann}, as illustrated in Figure \ref{fig:labeling}. Additionally, we included some non-smoke background images in our evaluation dataset to ensure robust testing.

We experimented with two state-of-the-art UDA approaches for semantic segmentation: AdaptSegNet \cite{tsai2020learning} and AdvEnt \cite{vu2019advent}. The details of these methods are discussed in Sections \ref{sec:asn} and \ref{sec:advent}. Our aim is to assess the extent to which these techniques can bridge the domain gap and improve segmentation performance on real-world smoke images.

\subsubsection{AdaptSegNet}
\label{sec:asn}
AdaptSegNet, proposed by Tsai et al. \cite{tsai2020learning} is an unsupervised domain adaptation (UDA) method which utilizes adversarial learning to adapt at output image space. 

\begin{figure}[htbp]
    \centering
    \includegraphics[width=\columnwidth]{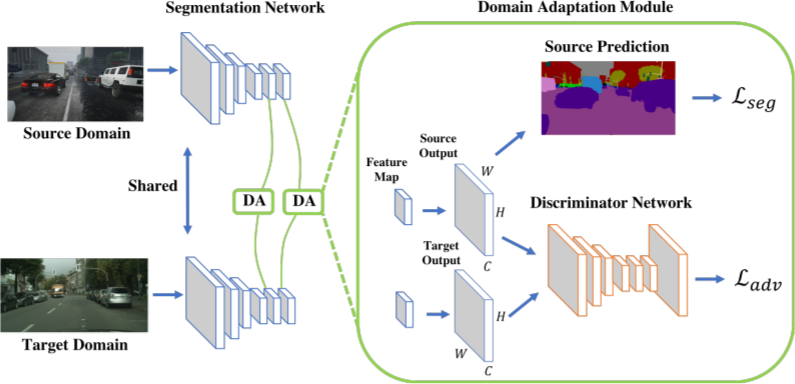}
    \caption{AdaptSegNet model architecture, adapted from \cite{tsai2020learning}}
    \label{fig:adaptsegnet_model}
\end{figure}

As illustrated in Figure \ref{fig:adaptsegnet_model}, the AdaptSegNet framework consists of two key components: a segmentation module and a domain discriminator. These components are trained jointly, with the segmentation module learning to generate accurate segmentation masks for the source domain using standard cross-entropy loss. Meanwhile, the domain discriminator attempts to distinguish between the segmentation outputs of the source and target domain images. An adversarial loss is applied to encourage the segmentation module to generate consistent outputs across both domains, effectively minimizing the domain discrepancy.

In our experiments, we utilized the multi-scale variant of AdaptSegNet to better capture features at different resolutions. We trained the model using our labeled synthetic smoke images as the source domain and the unlabeled real-world smoke images as the target domain. This setup aims to adapt the segmentation model to perform effectively on real-world smoke images despite the absence of labeled target data.

\subsubsection{AdvEnt}
\label{sec:advent}
AdvEnt, proposed by Vu et al. \cite{vu2019advent}, introduces an adversarial entropy minimization technique for domain adaptation, specifically targeting semantic segmentation tasks. The core idea of entropy minimization arises from the observation that models trained on synthetic data tend to produce low-entropy (high-confidence) predictions on synthetic images but generate high-entropy (low-confidence) predictions when applied to real-world images.

\begin{figure}[htbp]
    \centering
    \includegraphics[width=\columnwidth]{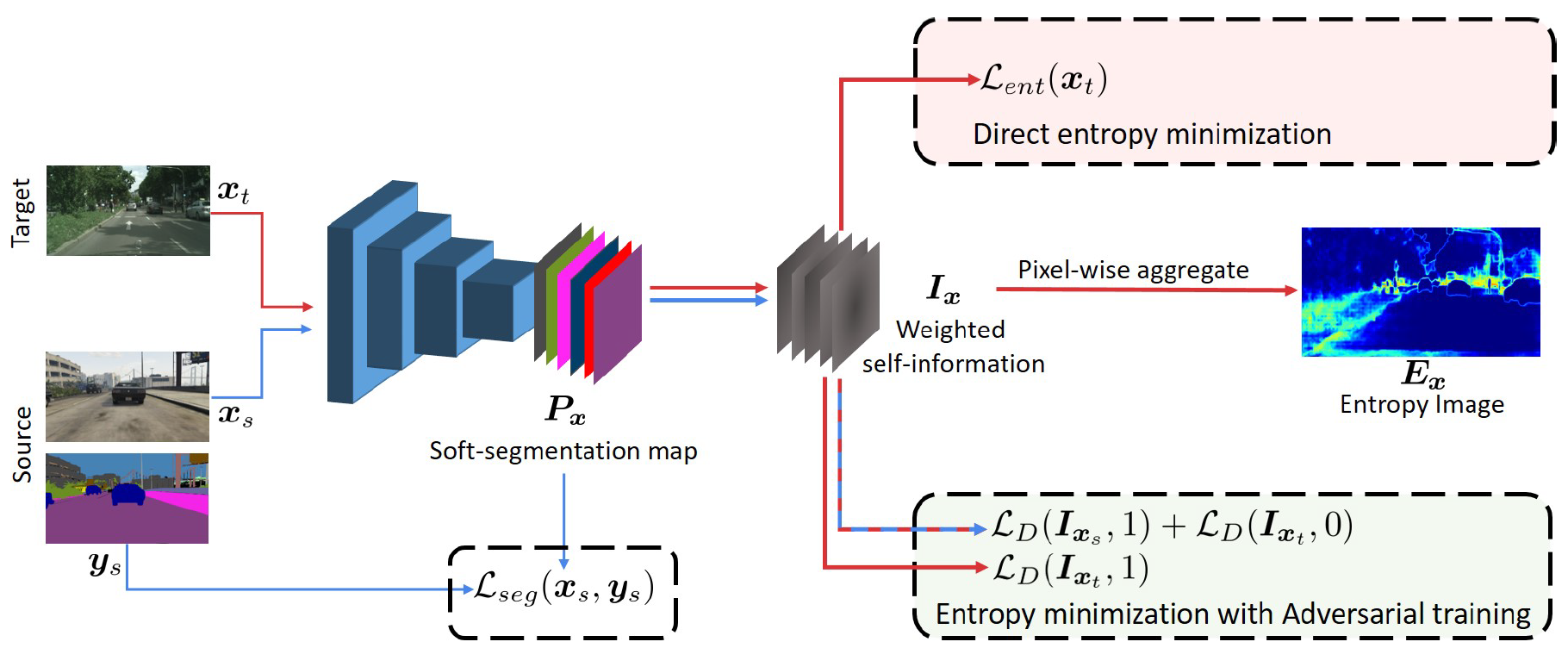}
    \caption{AdvEnt model architecture, taken from \cite{vu2019advent}}
    \label{fig:advent_model}
\end{figure}

In our experiments, we employed the adversarial framework illustrated in Figure \ref{fig:advent_model}. In this approach, the discriminator is trained to differentiate between the segmentation outputs of real and synthetic images, while the segmentation network is simultaneously trained to generate predictions that fool the discriminator. This adversarial interaction drives the segmentation network to minimize entropy on real-world images, thereby improving its performance by aligning the confidence levels of predictions across both domains.

\subsection{Bridging the Domain gap}
\label{sec:bridge}
Our experiments with Unsupervised Domain Adaptation (UDA) revealed that while segmentation models trained on synthetic data perform well on the source domain, they struggle to generalize to the target domain. Both the UDA and transfer learning results suggest that there is significant room for improvement.

We hypothesize that the domain gap between the synthetic and real-world data remains substantial, limiting the model’s performance. Specifically, the contrast between synthetic smoke and background is often much more distinct than in real-world scenarios, causing the network to perform accurate segmentation on synthetic images but only partial segmentation on real-world data. To address this challenge, we explored multiple techniques aimed at closing the domain gap, which are described in the following sections.

\subsubsection{Style Transfer}
\begin{figure}[htbp]
    \centering
    \includegraphics[width=\columnwidth]{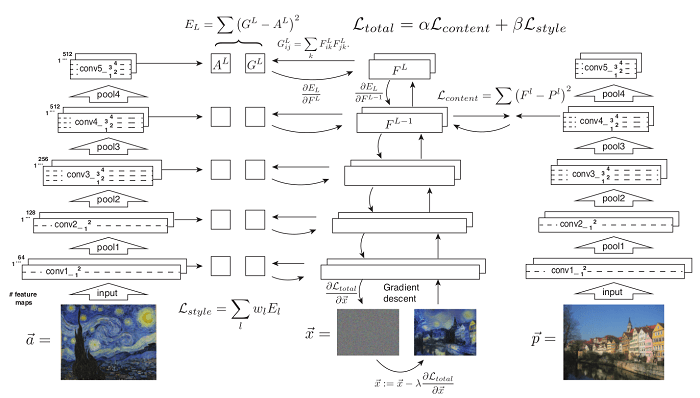}
    \caption{Neural style transfer, taken from \cite{gatys2015style}}
    \label{fig:neural_style_transfer}
\end{figure}

Neural style transfer is a technique that synthesizes an output image by combining the content of one image (the "content image") with the texture and style of another (the "style image"), as illustrated in Figure \ref{fig:neural_style_transfer}. For our experiment, the synthetic smoke image was used as the content, with the goal of applying the visual style of real smoke plumes. However, we aimed to limit the style transfer to the smoke plume regions alone.

To achieve this, we used a pretrained VGG19 network to extract feature maps from both the style and content images, with the content restricted to the smoke plume region through the use of label masks. We minimized both the style loss, which ensures the generated image captures the texture of the real smoke, and the content loss, which preserves the structure of the synthetic plume.

Despite the successful transfer of style to the masked smoke plume, the resulting stylized smoke struggled to blend naturally with the background, leading to visual artifacts. This limitation highlights a challenge when applying style transfer in a targeted manner achieving a seamless integration of the stylized foreground with the background remains difficult.

\subsubsection{Pix2Pix GAN}

The Pix2Pix GAN utilizes a PatchGAN discriminator that outputs a matrix of probabilities rather than a single scalar value. Each element in this matrix corresponds to a 70×70 patch of the input image and determines whether that patch is real or fake. The generator in Pix2Pix is based on the U-Net architecture, which uses skip connections to maintain spatial information and generate the segmented output map.

\begin{figure}[htbp]
    \centering
    \includegraphics[width=\columnwidth]{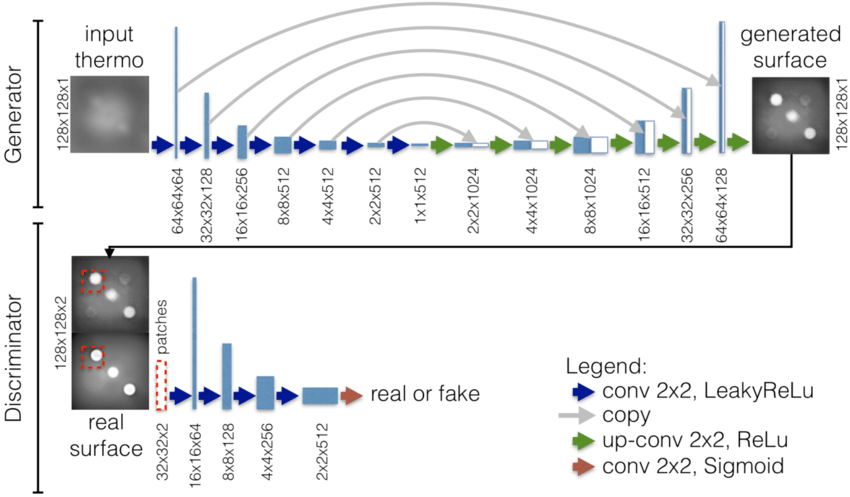}
    \caption{Pix2Pix GAN Generator architecture from \cite{pix2pix}}
    \label{fig:pix2pix}
\end{figure}

In our project, we employed the Pix2Pix GAN model (Figure \ref{fig:pix2pix}) for image-to-image translation to create more realistic smoke images. Specifically, we used real smoke images from the source domain and their corresponding binary masks from the target domain, generating additional binary masks in the process. By using binary masks as input from the source domain and real smoke images from the target domain, the model aimed to synthesize more realistic smoke images.

However, the generated outputs were hazy, with significant artifacts, likely because background features were also being transferred to the target domain along with the smoke features. As a result, these outputs were unsuitable for inclusion in the extended training set.

One major limitation of the Pix2Pix GAN is its reliance on paired datasets for image-to-image translation, meaning the source and target images must correspond to the same scene or location. This constraint proved problematic for our use case, as the real and synthetic smoke images did not always originate from identical locations. To overcome this limitation, we transitioned to CycleGAN, which enables unpaired image-to-image translation.

\subsubsection{Cycle GAN}
\begin{figure}[htbp]
    \centering
     \includegraphics[width=\columnwidth]{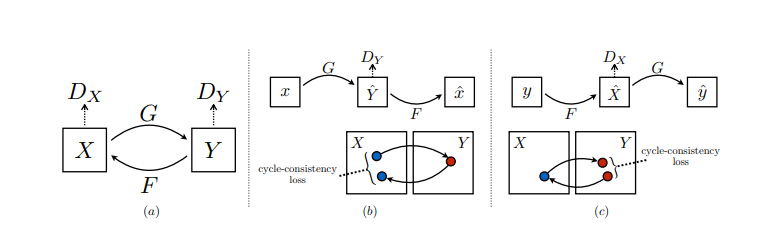}
    \caption{Cycle GAN Generator and Discriminator mappings from \cite{cycle_GAN}}
    \label{fig:cycleGAN}
\end{figure}

CycleGAN (Figure \ref{fig:cycleGAN}) extends the Pix2Pix architecture by incorporating two generator models and two discriminator models, allowing for simultaneous bidirectional translation between domains. Unlike Pix2Pix, CycleGAN supports unpaired datasets, meaning the source and target images do not need to correspond to the same scene or location. Additionally, the model can convert images in both directions—translating from the source domain to the target domain and vice versa—using the same architecture.

To regularize these translations, CycleGAN employs two cycle consistency losses. The forward consistency loss ensures that translating an image from one domain to the other and back again yields a result close to the original. Similarly, the backward consistency loss ensures the same consistency when translating in the opposite direction. These losses enforce stable and reliable mappings between the two domains.

In our first experiment, we treated real smoke images as the source domain and non-smoke images as the target domain to generate additional realistic smoke data. In the second experiment, we reversed the domains, using real smoke as the source domain and synthetic smoke as the target domain. This allowed us to test whether CycleGAN could generate more realistic synthetic smoke images that better align with the characteristics of real-world smoke.

\subsubsection{Image Matting}
\begin{figure}[htbp]
    \centering
    \includegraphics[width=\columnwidth]{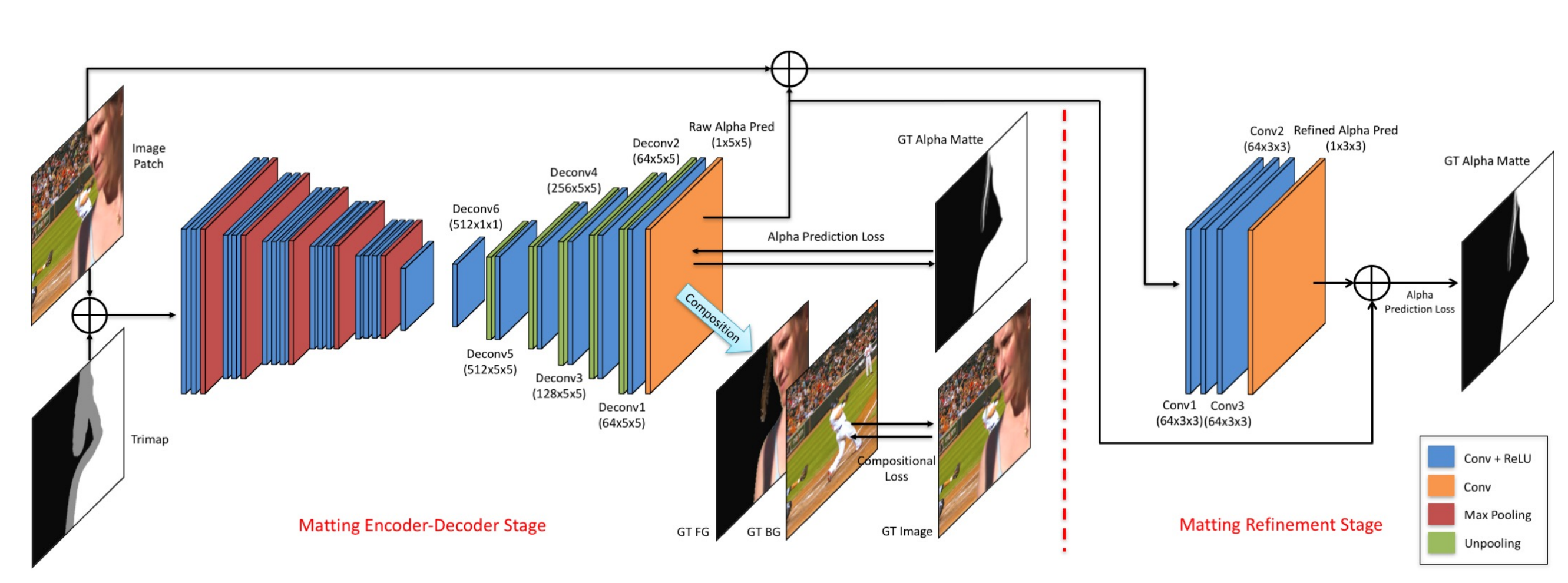}
    \caption{Deep Image Matting architecture, taken from \cite{xu2017matting}}
    \label{fig:deep_image_matte}
\end{figure}

To address the issue of sharp, unrealistic boundaries and poor blending of synthetic smoke into backgrounds, we explored deep image matting techniques \cite{xu2017matting}. Image matting involves a deep convolutional encoder-decoder network that takes an input image along with a corresponding trimap and predicts the alpha matte of the image (Figure \ref{fig:deep_image_matte}). A secondary, smaller convolutional network further refines the alpha matte to improve accuracy and sharpen the edges of the predicted mask.

The predicted alpha matte allows for precise separation of the foreground (smoke) from the background, enabling us to composite the smoke plume onto new background images seamlessly. We applied this technique to predict the alpha matte for real smoke plumes and blend them with non-smoke backgrounds to generate more realistic composite images.

However, one limitation of this approach is the need for manual annotation to generate trimaps for real smoke images, which is both time-consuming and resource-intensive. Due to these constraints, we were unable to create sufficient trimaps for real-world smoke images within the available time and resources. Despite these challenges, deep image matting shows promise for improving domain adaptation by creating more realistic training data through better foreground-background integration.

\section{Experiments}
\label{sec:exp}

\subsection{Transfer Learning}
\begin{figure}
    \centering
    \begin{subfigure}{.5\columnwidth}
        \includegraphics[width=\columnwidth, height = 0.4\columnwidth]{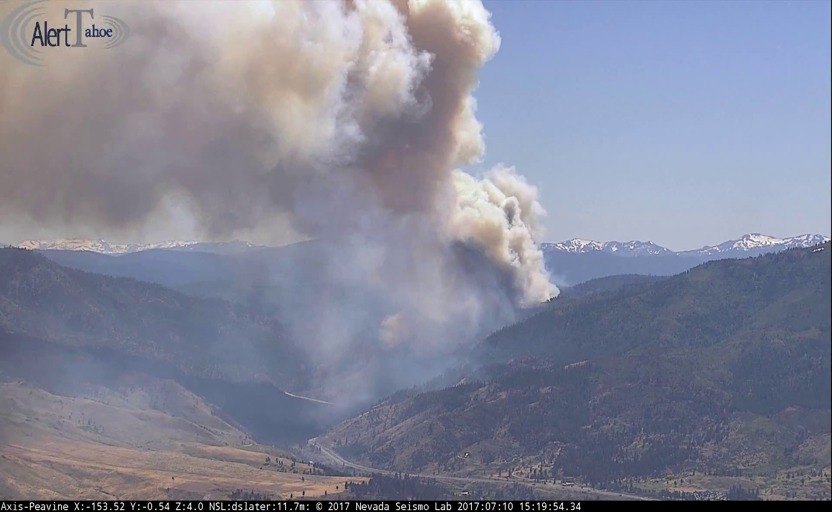}
    \end{subfigure}%
    ~
    \begin{subfigure}{.5\columnwidth}
        \includegraphics[width=\columnwidth, height = 0.4\columnwidth]{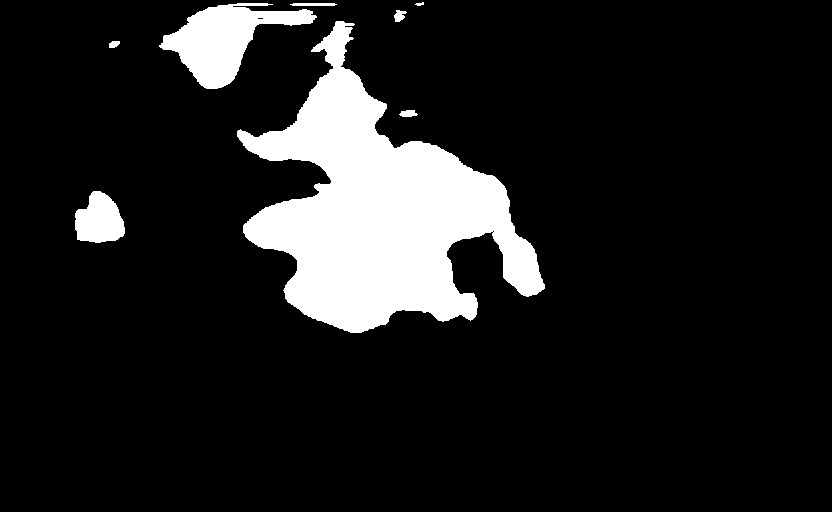}
    \end{subfigure}
    
    \begin{subfigure}{.5\columnwidth}
        \includegraphics[width=\columnwidth, height = 0.4\columnwidth]{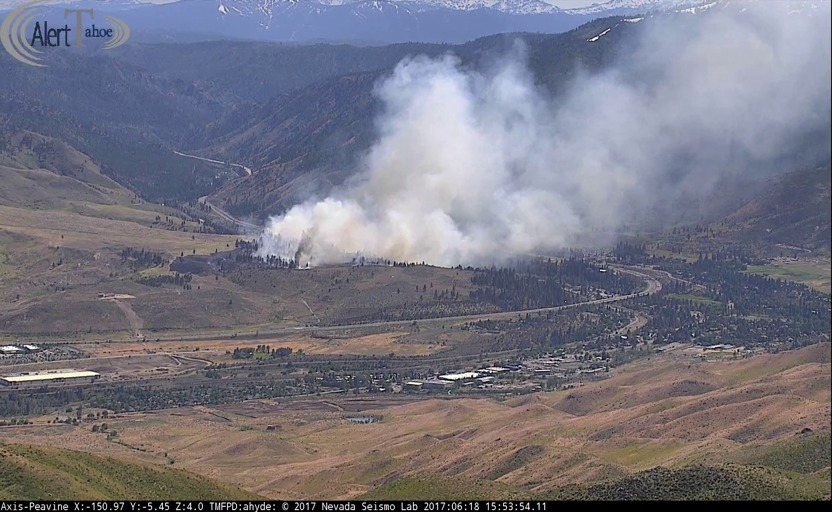}
    \end{subfigure}%
    ~
    \begin{subfigure}{.5\columnwidth}
        \includegraphics[width=\columnwidth, height = 0.4\columnwidth]{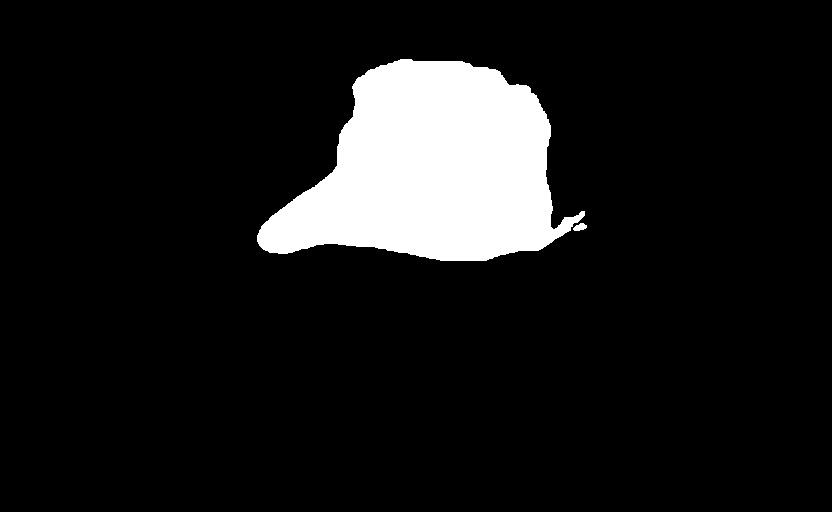}
    \end{subfigure}
    \caption{Sample images from the unlabelled target dataset and the model's corresponding predictions.}
    \label{fig:tf_output}
\end{figure}

We fine-tune the U-Net model, initially trained on the source data, using 80\% of the labeled target training data. The remaining 20\% of the labeled target data is used for evaluation. Additionally, we present some qualitative results on the unlabelled target data in Figure \ref{fig:tf_output}, showcasing the model's ability to generalize. The quantitative performance, measured using the mean Intersection over Union (mIoU), is reported in Table \ref{tab:miou} for the remaining labeled target data. This evaluation highlights the model’s effectiveness after fine-tuning on a limited subset of the target data.

\subsection{Unsupervised Domain Adaptation}
This section presents the results of state-of-the-art unsupervised domain adaptation (UDA) models, which we adapted to our datasets and evaluated through experiments.

\subsubsection{AdaptSegNet}
We trained the multi-scale version of AdaptSegNet, using Cross-Entropy loss for segmentation and per-pixel Binary Cross-Entropy (BCE) loss for domain discrimination and adversarial loss. The architecture configuration follows the original design by the authors, and we trained the model from scratch using our dataset. Hyperparameters were tuned to balance the various losses, with a specific focus on the weight for the adversarial loss. After experimentation, we set the adversarial loss weight to $\lambda_{adv} = 0.1$, while keeping other hyperparameters fixed.

\begin{figure}
    \centering
    \begin{subfigure}{.5\columnwidth}
        \includegraphics[width=\columnwidth, height = 0.4\columnwidth]{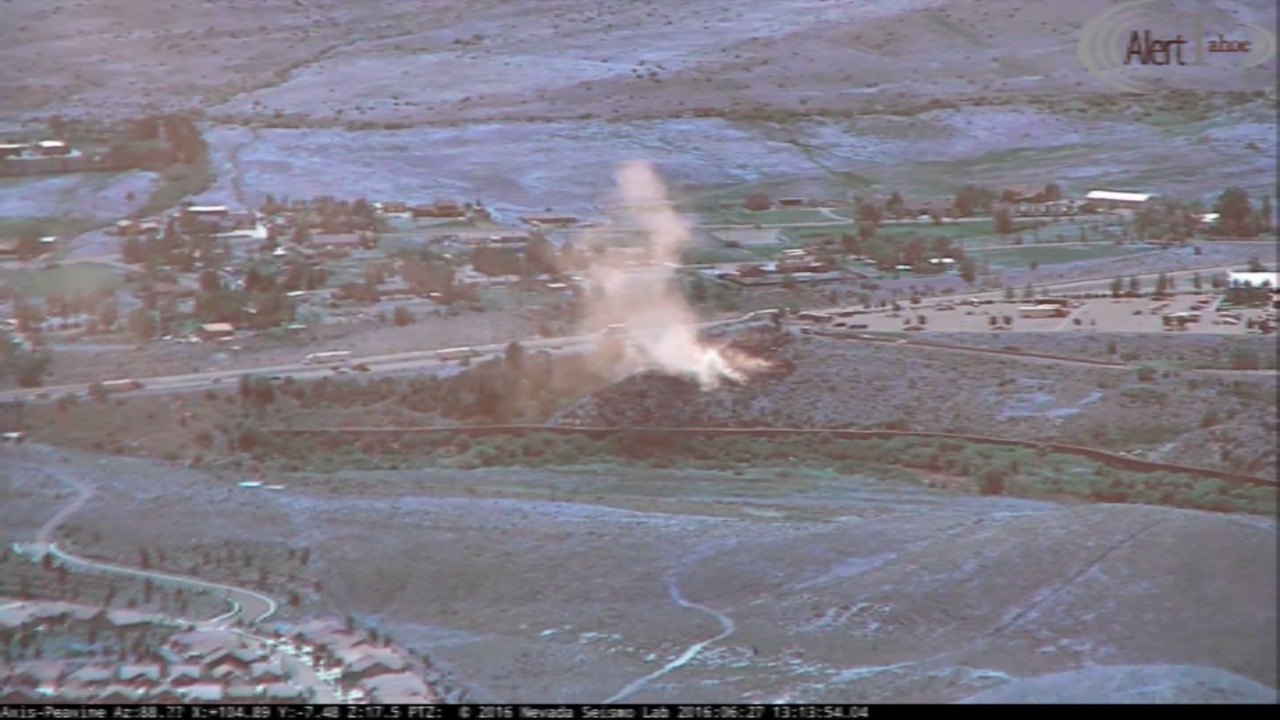}
    \end{subfigure}%
    ~
    \begin{subfigure}{.5\columnwidth}
        \includegraphics[width=\columnwidth, height = 0.4\columnwidth]{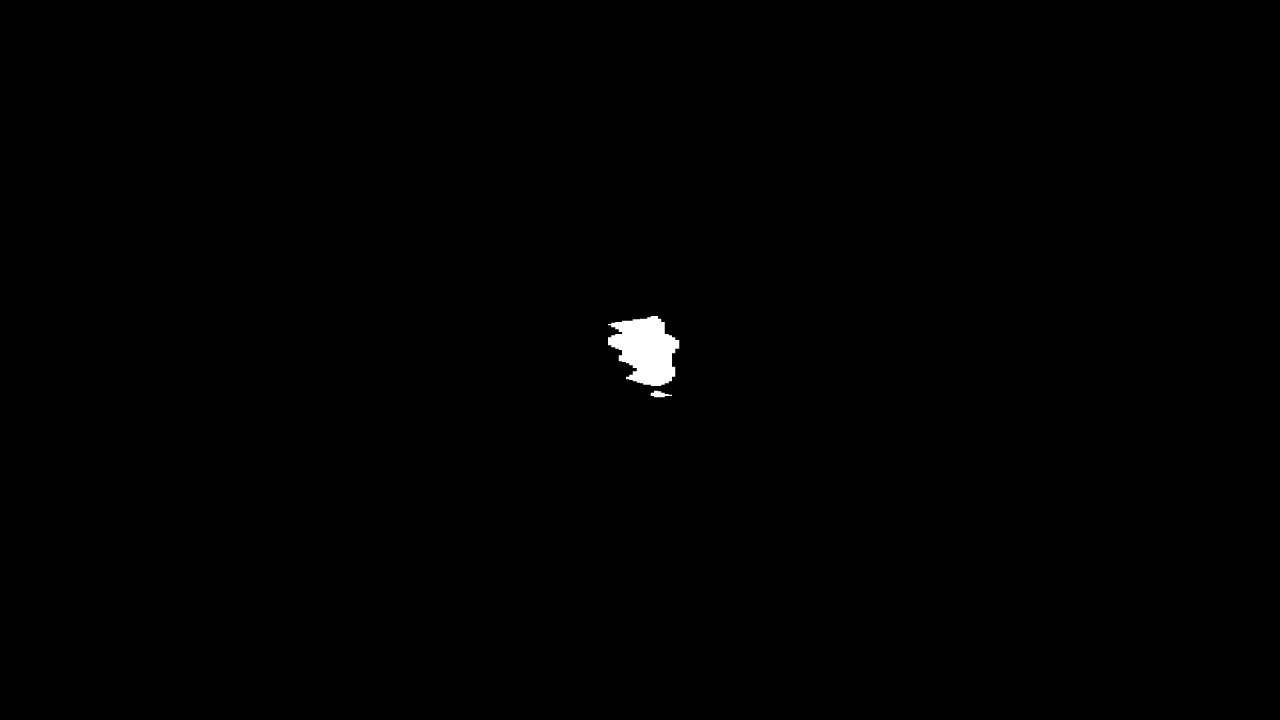}
    \end{subfigure}
    
    \begin{subfigure}{.5\columnwidth}
        \includegraphics[width=\columnwidth, height = 0.4\columnwidth]{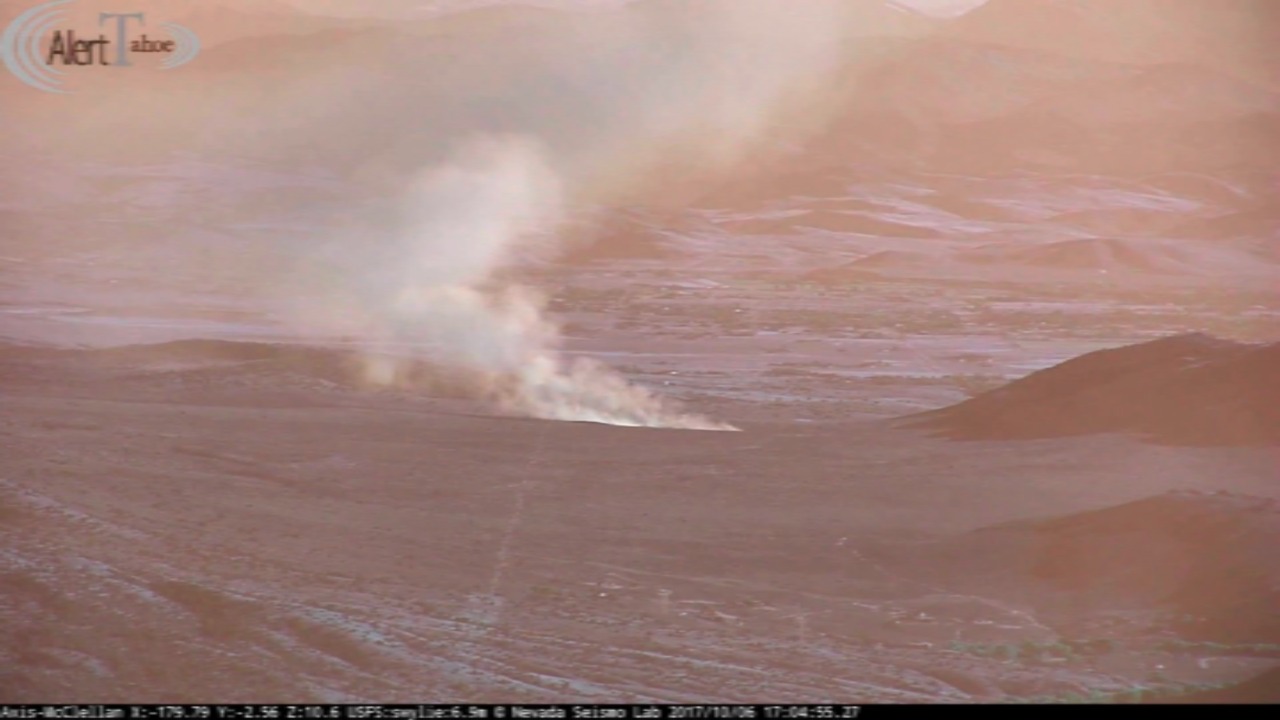}
    \end{subfigure}%
    ~
    \begin{subfigure}{.5\columnwidth}
        \includegraphics[width=\columnwidth, height = 0.4\columnwidth]{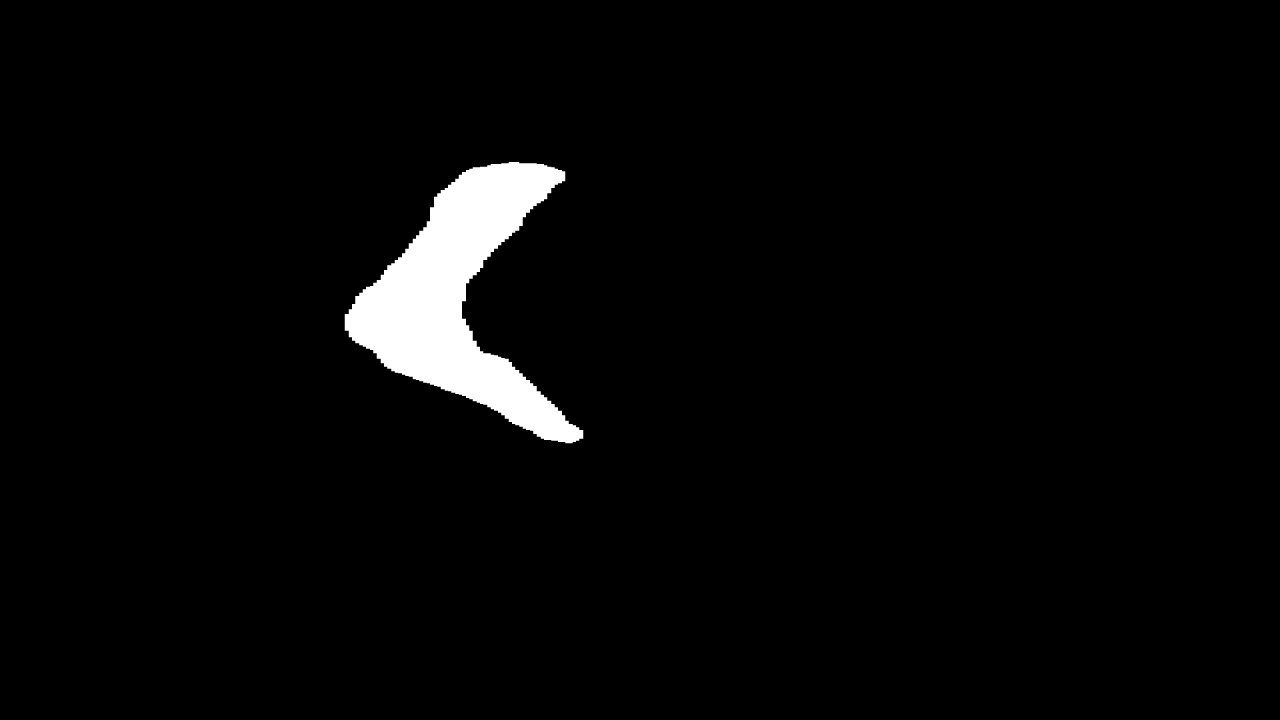}
    \end{subfigure}

    \caption{Sample images from the target dataset and the corresponding outputs generated by the AdaptSegNet model.}
    \label{fig:adaptsegnet_qualitative_outputs}
\end{figure}

Figure \ref{fig:adaptsegnet_qualitative_outputs} shows qualitative results on target data, and Table \ref{tab:miou} provides quantitative results in terms of Mean Intersection over Union (mIoU) on labeled target data.

\subsubsection{AdvEnt}
\begin{figure}[ht]
\centering
\begin{subfigure}{.5\columnwidth}
    \centering
    \includegraphics[width=0.9\linewidth]{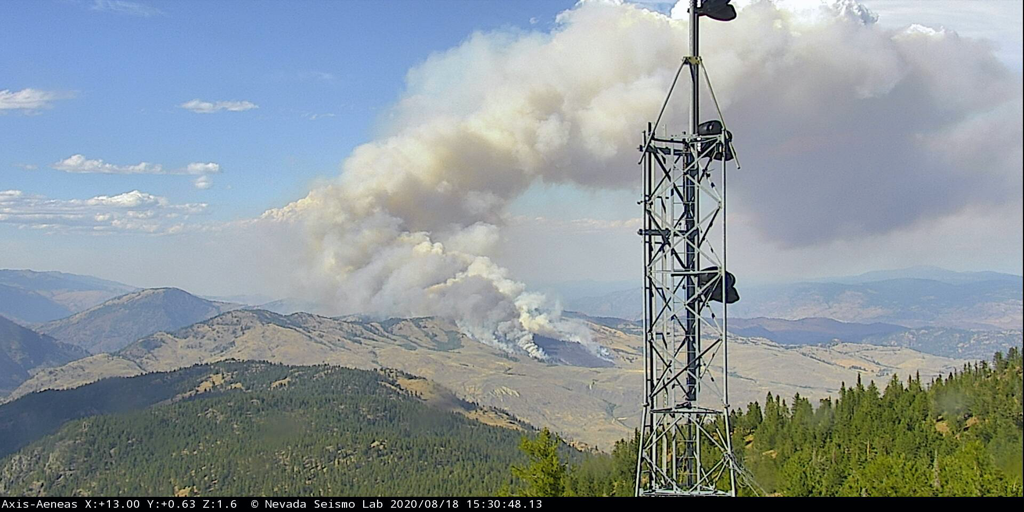}
    \label{advent_res_a}
    \caption{}
\end{subfigure}%
\hfill
\begin{subfigure}{.5\columnwidth}
    \centering
    \includegraphics[width=0.9\linewidth]{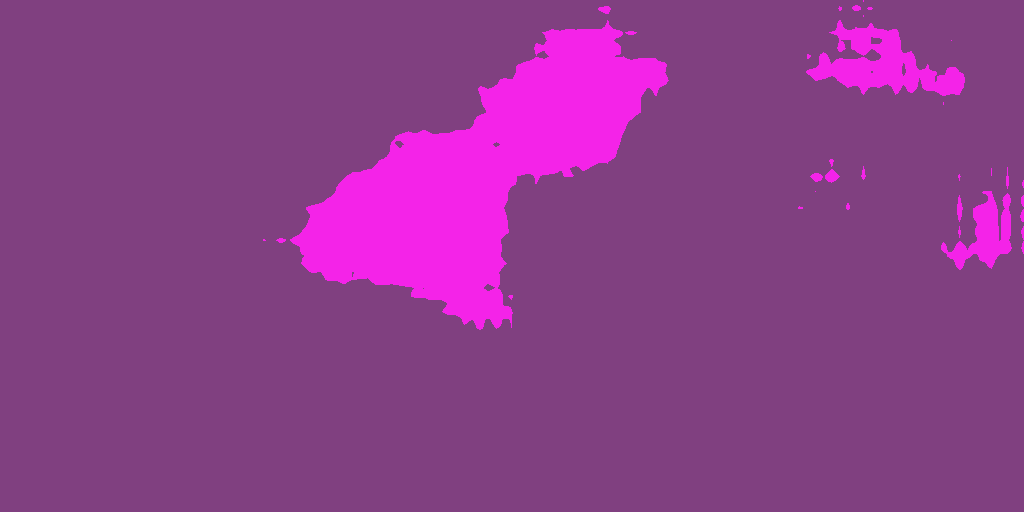}
    \label{adevent_res_b}
    \caption{}
\end{subfigure}%
% \hfill
\begin{subfigure}{.5\columnwidth}
    \centering
    \includegraphics[width=0.9\linewidth]{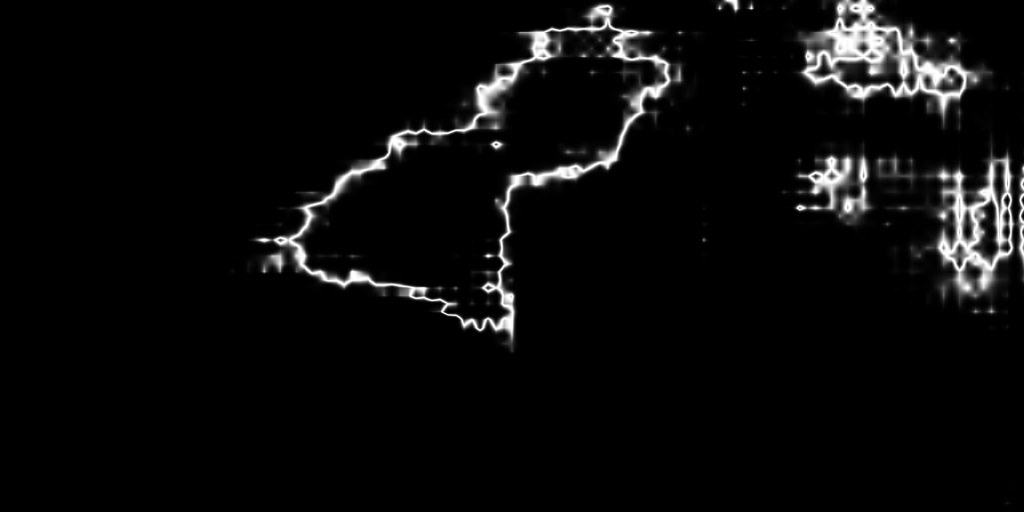}
    \label{advent_res_c}
    \caption{}
\end{subfigure}%

\caption{AdvEnt minimizes entropy for unsupervised domain adaptation through an adversarial approach. The model takes a real-world smoke image (target) as input (a), generating a binary label mask (b) and an entropy mask (c) for the smoke.}
\label{fig:advent_res}
\end{figure}

We implemented the AdvEnt model using a DeepLabv2 backbone. The model was trained for 12,000 iterations using different weights for the adversarial loss term to achieve optimal performance. Figure \ref{fig:advent_res} presents the qualitative results on real-world target images used for evaluation, while Table \ref{tab:miou} summarizes the quantitative results using mIoU.

\begin{table}[htbp]
    \centering
    \begin{tabular}{|p{0.12\linewidth}|p{0.30\linewidth}|p{0.24\linewidth}|p{0.15\linewidth}|}
            \hline
         Model & UNet (Transfer learning) & AdaptSegNet & AdvEnt \\
         \hline
         mIoU & 19.24 & 6.75 & 3.74\\
         \hline
    \end{tabular}
    \caption{Mean IoU on the labelled target dataset}
    \label{tab:miou}
\end{table}

\subsection{Bridging the Domain gap}
Our UDA experiments reveal that models trained on synthetic data struggle to generalize to real-world smoke due to significant domain discrepancies. The primary challenge lies in the stark contrast between synthetic smoke and its background, which differs from the subtler visual characteristics of real smoke. Additionally, the amorphous, diffuse nature of smoke makes segmentation more challenging than for objects with well-defined boundaries.

\subsubsection{Style Transfer}
\begin{figure}[ht]
\centering
\begin{subfigure}{.5\columnwidth}
    \centering
    \includegraphics[width=0.9\linewidth]{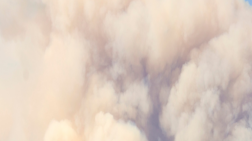}
    % \label{advent_res_a}
    \caption{}
\end{subfigure}%
% \hfill
\begin{subfigure}{.5\columnwidth}
    \centering
    \includegraphics[width=0.9\linewidth]{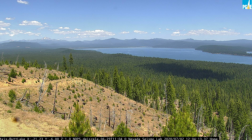}
    % \label{adevent_res_b}
    \caption{}
\end{subfigure}%
\hfill
\begin{subfigure}{.5\columnwidth}
    \centering
    \includegraphics[width=0.9\linewidth]{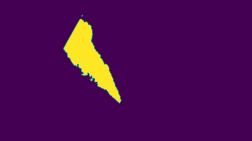}
    % \label{advent_res_c}
    \caption{}
\end{subfigure}%
\begin{subfigure}{.5\columnwidth}
    \centering
    \includegraphics[width=0.9\linewidth]{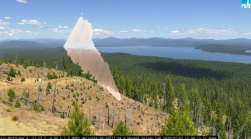}
    % \label{advent_res_c}
    \caption{}
\end{subfigure}%

\caption{a) shows the style image with real smoke texture, b) the content image requiring a synthetic smoke plume, c) the mask constraining the plume, and d) the resulting synthetic image.}
\label{fig:style_transfer_result}
\end{figure}

As shown in Figure \ref{fig:style_transfer_result}, the stylized smoke retains sharp boundaries and exhibits a different color scheme and brightness compared to the background, resulting in poor blending. These differences prevent the stylized output from resembling natural fire smoke. This highlights the persistent domain gap, as realistic segmentation becomes challenging when the model encounters the fuzzy, irregular boundaries typical of smoke plumes. While segmentation models can easily detect objects with clear edges, handling amorphous structures like smoke proves more difficult, further complicating domain adaptation efforts.

% \begin{figure}[ht]
% \includegraphics[width=\linewidth]{figures/style_img.png}
% \caption{On the top left is the style image, containing the real smoke's texture and style, and on the top right is the content image, which requires the addition of a synthetic smoke plume. The bottom left shows the mask used to constrain the smoke plume, and the bottom right displays the resulting synthetic image.}
% \label{fig:style_transfer_result}
% \end{figure}

\subsubsection{Pix2pix GAN}
The Pix2Pix GAN was utilized for pixel-level image-to-image translation by leveraging real smoke images and their corresponding binary masks to generate synthetic smoke data. This approach allowed the GAN to learn the mapping between the binary masks and real smoke patterns, producing realistic smoke textures in the synthetic domain.

\begin{figure}[htbp]
    \centering
    \includegraphics[width=\linewidth, height = 0.6\linewidth]{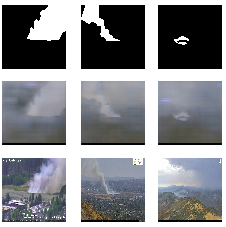}
    \caption{Results obtained when we used real smoke images (third row) and their corresponding masks (first row) to obtain more real smoke images(second row) and to reduce the domain shift}
    \label{fig:pix2pix_res}
\end{figure}

As illustrated in Figure \ref{fig:pix2pix_res}, the second row shows examples of the synthetic smoke data generated by the model. While this method offers a straightforward way to synthesize new smoke images, ensuring the generated smoke closely mimics real-world smoke remains a challenge.

\subsubsection{CycleGAN}
We explored two approaches using CycleGAN for domain translation:
\begin{enumerate}[label=\alph*)]
    \item Translation from the real-world smoke image domain to the real-world non-smoke image domain.
    \item Translation from the real-world smoke image domain to the synthetic smoke image domain.     
\end{enumerate}

\begin{figure}[ht]
\centering
\begin{subfigure}{.5\columnwidth}
    \centering
    \includegraphics[width=0.9\columnwidth, height=0.5\columnwidth]{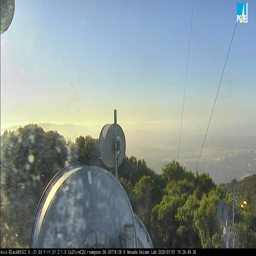}
    \label{matting_res_a}
    \caption{}
\end{subfigure}%
\begin{subfigure}{.5\columnwidth}
    \centering
    \includegraphics[width=0.9\columnwidth, height=0.5\columnwidth]{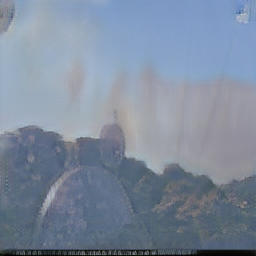}
    \label{matting_res_b}
    \caption{}
\end{subfigure}%

\begin{subfigure}{.5\columnwidth}
    \centering
    \includegraphics[width=0.9\columnwidth, height=0.5\columnwidth]{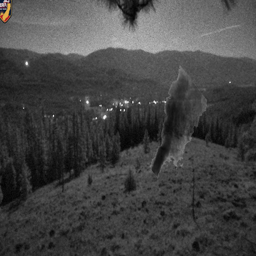}
    \label{matting_res_c}
    \caption{}
\end{subfigure}%
\begin{subfigure}{.5\columnwidth}
    \centering
    \includegraphics[width=0.9\columnwidth, height=0.5\columnwidth]{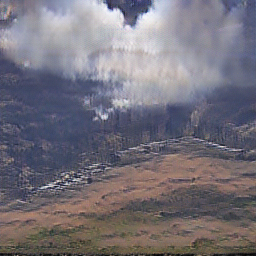}
    \label{cyclegan_res_d}
    \caption{}
\end{subfigure}%
\caption{CycleGAN translating a real-world non-smoke domain image (a) to a smoke domain image (b), and a synthetically generated smoke domain image (c) to a real-world smoke domain image (d). The outputs demonstrate artifacts, reduced sharpness, and altered colors, highlighting the limitations of CycleGAN for realistic smoke generation.} 
\label{fig:cyclegan_res}
\end{figure}

The results of both approaches are shown in Figure \ref{fig:cyclegan_res}. However, the generated outputs demonstrate several limitations. The smoke synthesized by CycleGAN fails to accurately resemble real-world smoke and introduces unwanted artifacts into the images. Additionally, the overall image quality deteriorates, making it unsuitable for practical applications. These issues suggest that CycleGAN, in its current form, is not a viable solution for generating realistic smoke imagery for use in real-world settings.

\subsubsection{Image Matting}
\label{sec:img_matting}

\begin{figure}[ht]
\centering
\begin{subfigure}{.5\columnwidth}
    \centering
    \includegraphics[width=0.9\columnwidth, height = 0.5\columnwidth]{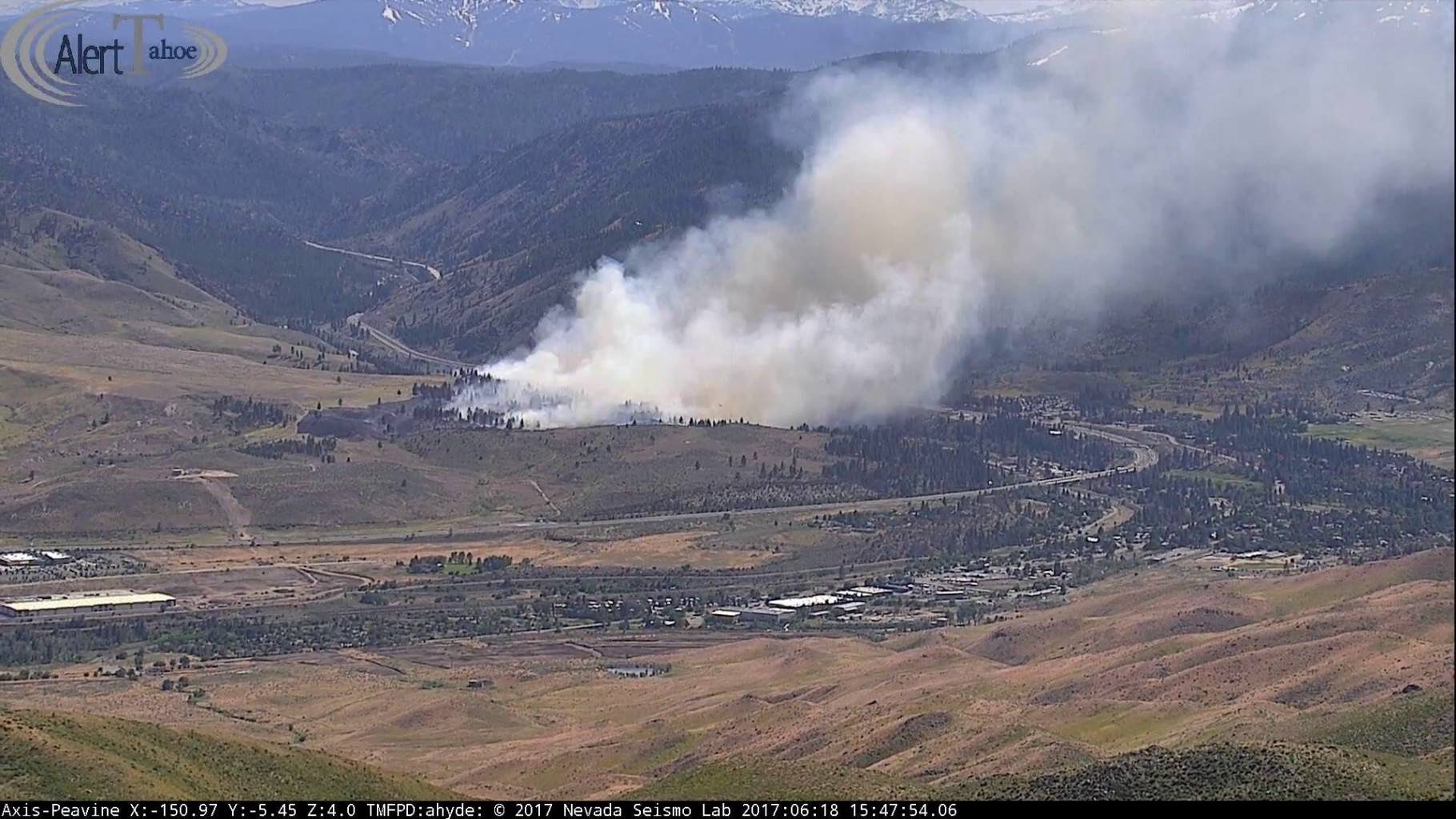}
    \caption{}
\end{subfigure}%
\begin{subfigure}{.5\columnwidth}
    \centering
    \includegraphics[width=0.9\columnwidth, height = 0.5\columnwidth]{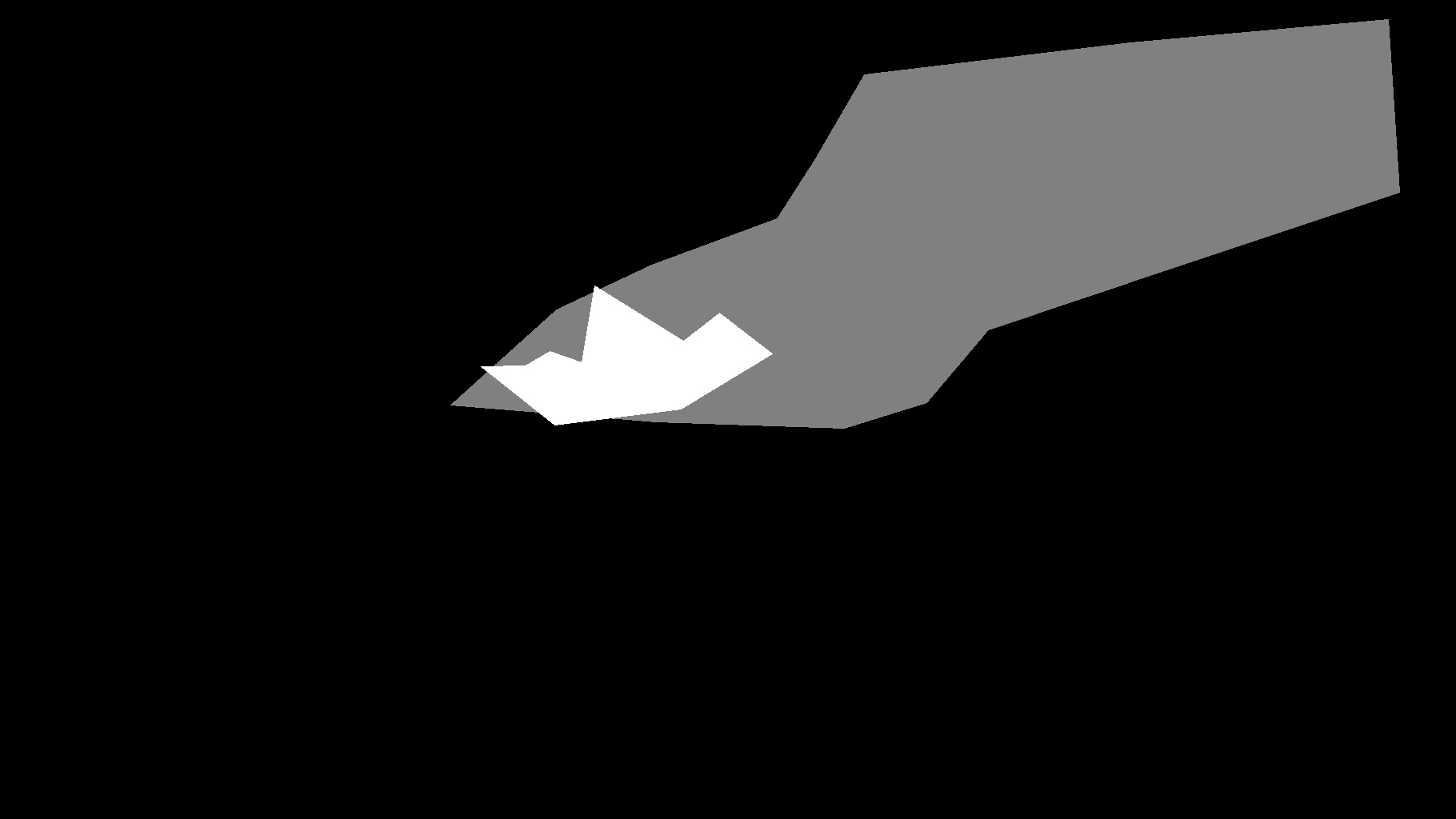}
    \caption{}
\end{subfigure}%

\begin{subfigure}{.5\columnwidth}
    \centering
    \includegraphics[width=0.9\columnwidth, height = 0.5\columnwidth]{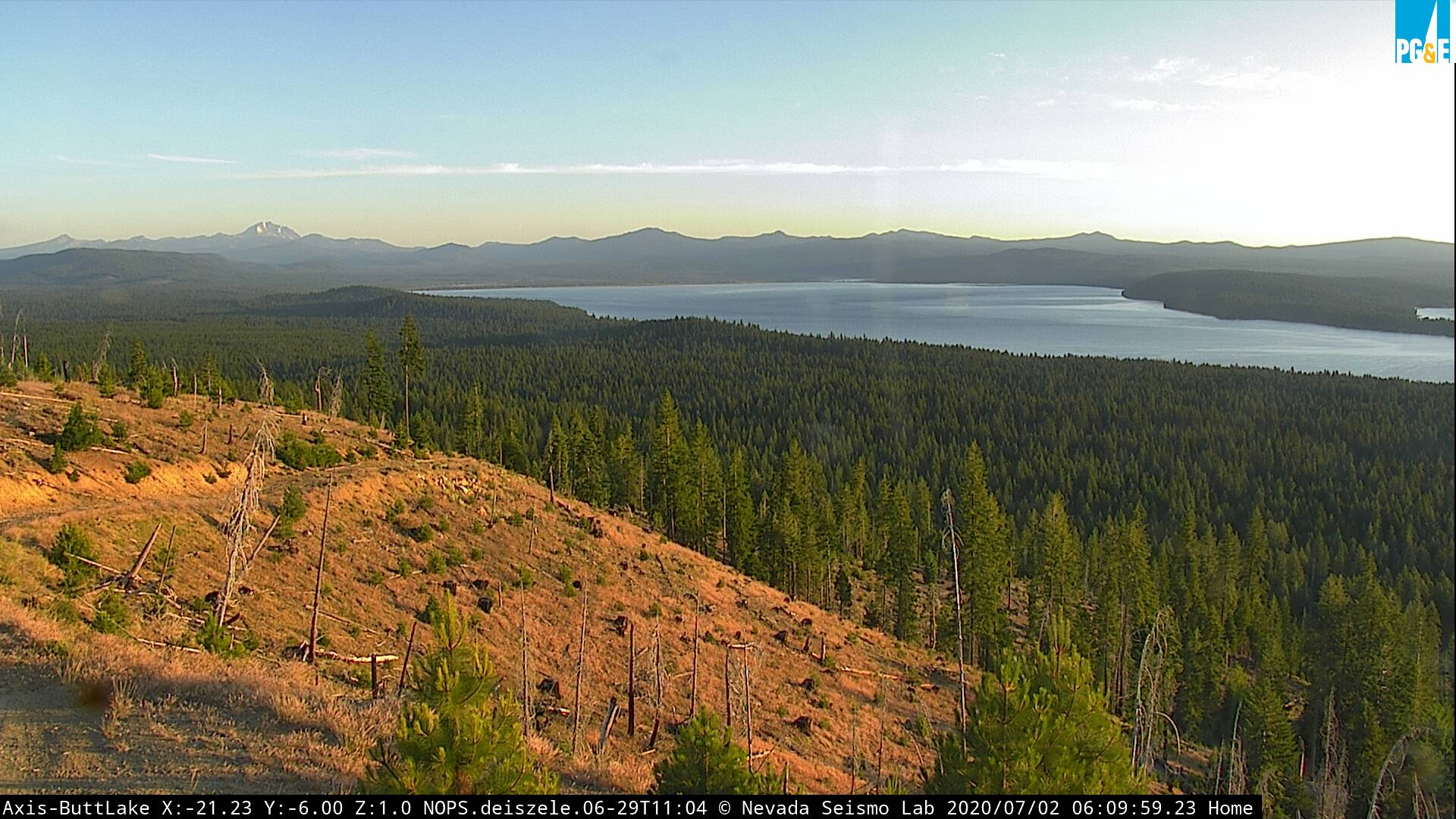}
    \caption{}
\end{subfigure}%
\begin{subfigure}{.5\columnwidth}
    \centering
    \includegraphics[width=0.9\columnwidth, height = 0.5\columnwidth]{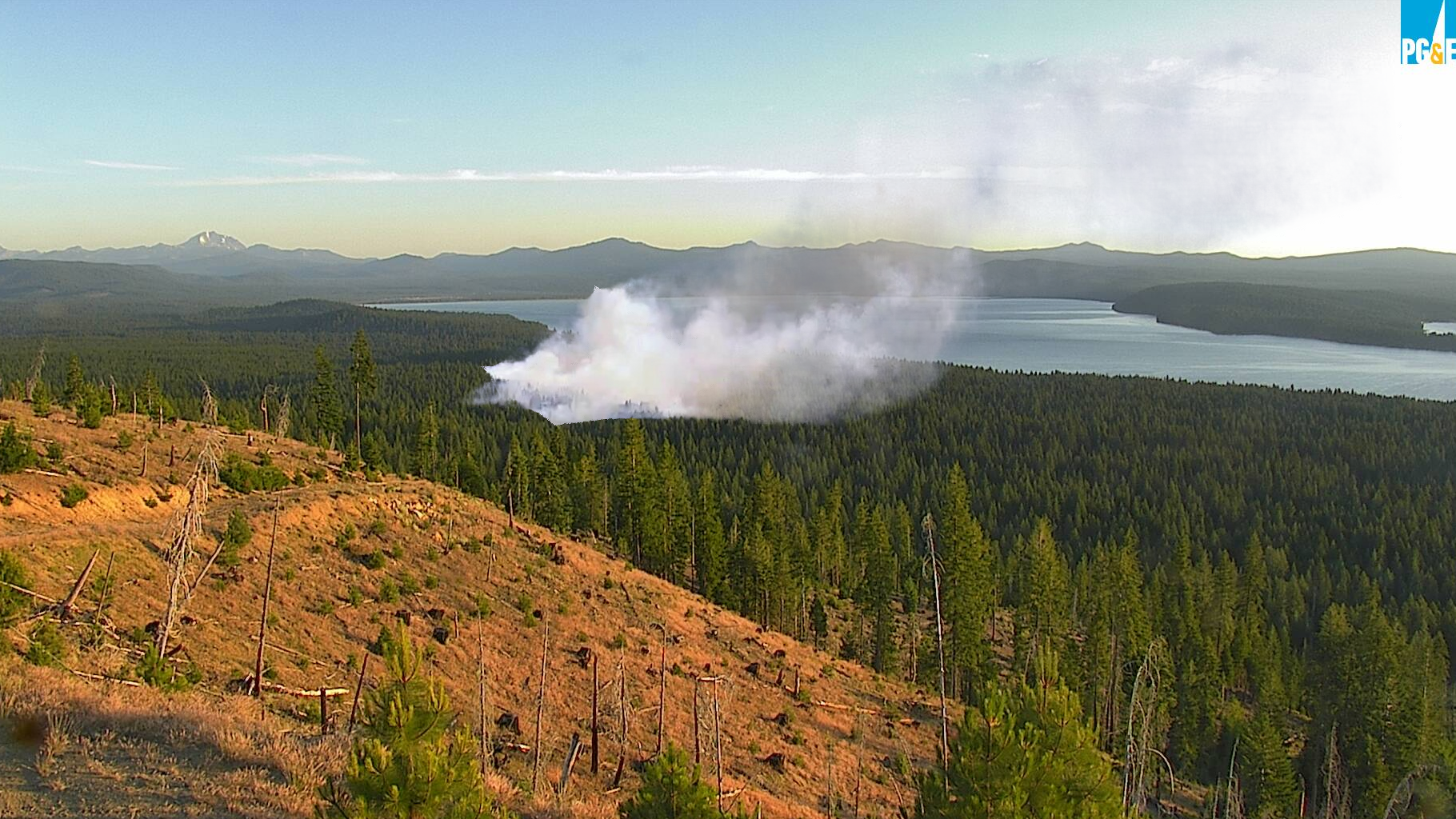}
    \caption{}
\end{subfigure}%
\caption{Deep Image Matting takes as input a real smoke image (a) and a corresponding trimap (b) to generate a refined alpha matte. The smoke foreground is then composited with a new background (c) to create a synthetic image (d). In the trimap, the white region indicates a definite mask, while the gray region represents an uncertain or probable mask.}

\label{fig:image_matting}
\end{figure}

The results of applying Deep Image Matting to composite real-world smoke onto a non-smoke background are presented in Figure \ref{fig:image_matting}. For this experiment, the input trimap was manually generated using the VGG Annotator tool \cite{vgg_ann}. Although the results appear promising in reducing the domain gap, the method is not practical for large-scale use.

Generating accurate trimaps for multiple real-world images would be a labor-intensive process, and we lacked sufficient time to develop an automated approach. A potential solution could involve using synthetic white-background smoke images from our original synthetic dataset to automate trimap generation. However, due to resource constraints, we were unable to implement and evaluate this approach within the scope of this work.

\section{Conclusion and Future work}
In this work, we conducted experiments to detect and localize wildfire smoke using images collected from the Alert Wildfire camera network. Our objective was to explore the potential and limitations of employing unsupervised domain adaptation (UDA) techniques to perform semantic segmentation on real-world data, leveraging features learned from easily synthesized labeled synthetic datasets. This approach aims to minimize the need for labor-intensive manual annotation of real-world images.

We experimented with several state-of-the-art UDA techniques, and the results indicate a substantial domain gap between synthetic and real-world data, which limits segmentation performance. To address this gap, we explored advanced methods, including style transfer, CycleGAN, and Pix2Pix GAN. However, these techniques were unable to fully close the domain gap, as the generated outputs often exhibited artifacts, poor blending, or degraded image quality.

Among the explored methods, deep image matting offers a promising direction. If the process of generating trimaps can be automated—potentially by leveraging synthetic smoke images with white backgrounds—this could improve the quality of synthetic datasets, enabling more effective domain adaptation. Another avenue for future work involves exploring semi-supervised domain adaptation techniques, which combine limited labeled data with unlabeled data to improve model performance. Although semi-supervised methods were beyond the scope of this project, they represent a valuable direction for further research.

% {\small
% \bibliographystyle{ieee_fullname}
% \bibliography{reference.bib}
% }

\bibliography{main}

\end{document}